# Towards Integrated Rock Support Visualisation in 3D Point Cloud of Underground Mines

Dibyayan Patra[1], Simit Raval[1,*], Pasindu Ranasinghe[1], Bikram Banerjee[2], Ismet Canbulat[1]
[1]School of Minerals and Energy Resources Engineering, University of New South Wales, Sydney, NSW, Australia.
[2]School of Surveying and Built Environment, University of Southern Queensland, Toowoomba, QLD, Australia.
[*]Corresponding Author Email: simit@unsw.edu.au

**Abstract:** The effectiveness of rock support in underground mines depends on the interaction between installed rock bolts and the structural fabric of the surrounding rock mass. However, discontinuity characterisation and rock bolt identification are commonly treated as separate tasks, limiting their value for integrated support assessment. This study presents an automated framework for integrated rock support visualisation using 3D point clouds of underground mine excavations. The framework integrates structure mapping, rock bolt identification, discontinuity plane fitting, and bolt orientation estimation into a unified workflow optimised for accuracy and computational efficiency. The outputs are used to generate an integrated 3D visualisation of fitted discontinuity planes and bolt vectors, enabling direct assessment of their spatial intersections and geometric relationships. A complementary stereographic analysis of discontinuity poles and bolt orientations is also performed to evaluate overall bolting geometric effectiveness relative to the mapped structural fabric. Additionally, bolt-level quality metrics, including exposed protrusion length and deviation from the local roof normal, are visualised to support assessment of installation quality. The proposed framework is demonstrated on real underground metal mine scans, producing accurate structure mapping and rock bolt identification results in medium-scale point clouds. Overall, the study provides a practical step towards automated, integrated geotechnical assessment of rock support effectiveness without requiring manual measurements or additional in-situ data acquisition.

## 1. Introduction

The structural integrity of underground excavations is critically dependent on the effectiveness of the installed rock support system. In underground mining environments, rock bolts serve as the primary mechanical support element, anchoring potentially unstable rock mass to the surrounding competent rock and restraining failure mechanisms such as wedge and planar sliding, flexural and block toppling, and roof collapse [1]. Supplementary support elements, such as shotcrete and mesh, provide surface-level protection against small rock fragment detachment, but rock bolts constitute the principal structural mechanism for preventing large-scale rock mass failures. The design and quality of rock bolt installations are therefore directly governed by the structural geology of the surrounding rock mass, specifically the orientation, spacing, and persistence of discontinuity planes such as joints, faults, and bedding surfaces [2-6]. Discontinuities sharing similar characteristics and orientation are grouped into discontinuity sets, which represent the dominant structural fabric of the rock mass and define the principal planes of weakness governing its mechanical behaviour [7-10]. These discontinuities are formed by geological deformation in the Earth's crust and are exposed at the excavation boundary following underground mining. Rock bolts are installed across these discontinuities to provide tensile resistance against in-situ stress conditions, physically restraining the rock mass and preventing hazardous displacements and rockfalls. A bolt installed at a suboptimal orientation relative to a critical discontinuity plane, or in a region where significant structural features remain unanchored, may provide substantially reduced support effectiveness and represent a latent ground control hazard [1, 2].

This well-established relationship between rock mass structure and support design is routinely assessed by geotechnical engineers as part of standard underground mine monitoring workflows for stability analysis and predictive maintenance. However, current practice relies heavily on manual inspection where engineers physically assess exposed tunnel surfaces under low-light, hazardous conditions, recording structural orientations through compass-clinometer measurements and evaluating bolt installations through visual examination [8, 11-14]. These approaches are time-consuming, inherently subjective, prone to human error, and fundamentally limited in their spatial coverage and repeatability. Moreover, as mine operations increasingly extend into greater depths and more geologically complex environments, the inadequacy of manual monitoring methods presents a growing operational and safety concern.

The rapid advancement of remote sensing and laser scanning technologies for underground mines has created new opportunities for automated, high-resolution characterisation of underground mine environments [12, 15-17]. These systems capture dense 3D point clouds of tunnel surfaces at centimetre to millimetre resolution, encoding both the macro-scale geometry of the excavation and fine-scale features such as discontinuity surfaces and rock bolt protrusions. This geometric richness has motivated a growing body of research into automated structure mapping and rock bolt identification from point cloud data, with each problem having been studied independently in the literature. Structure mapping techniques have been developed to automatically characterise discontinuity sets and estimate their orientations from point cloud data, while separate approaches have been proposed for the automated detection and geometric characterisation of rock bolt protrusions as has been described in Section 2 in detail. However, despite significant progress in both areas independently, no study to date has brought the two together in an integrated framework utilising laser scanning. The co-registration of rock bolt orientations with the structural fabric of the surrounding rock mass, a fundamental requirement for meaningful support geometric effectiveness assessment, has not been addressed in the literature. This represents a critical gap, as the geotechnical value of either output is substantially enhanced when considered in the context of the other. A bolt orientation vector carries limited geotechnical meaning without knowledge of the discontinuity planes it is intended to intercept, and a characterised discontinuity plane provides limited actionable insight without knowledge of the support installed relative to it.

This study, for the first time, presents an automated integrated rock support visualisation framework that integrates and correlates discontinuity characterisation with rock bolt identification directly from underground mine point cloud data. The proposed framework processes point cloud data captured from underground metal mine tunnels through two complementary processing pipelines, a structure mapping workflow and a rock bolt identification workflow. Deliberate efforts were made to improve computational efficiency by reusing shared preprocessing modules and features across both workflows, reducing redundant calculations and lowering overall computational complexity without sacrificing robustness or accuracy. The framework further incorporates discontinuity plane fitting and rock bolt orientation estimation steps to convert the characterised structural and support features into explicit 3D geometric representations. These outputs are then integrated into two complementary visualisation representations forming the integrated rock support visualisation, an integrated 3D visualisation and spatial overlay of fitted discontinuity planes and intersecting bolt orientation vectors, and a joint stereographic projection of discontinuity poles and bolt orientations on a lower hemisphere stereonet. Together with computed bolt-level quality metrics, these representations provide a comprehensive, automated, and surface-faithful framework for geotechnical assessment of rock support effectiveness, requiring no manual measurements, no additional in-situ data capture, and no subsurface geometric assumptions.

## 2. Related Works

### 2.1. Structure Mapping Techniques

With the advent of laser scanning in mining and geology, structure mapping techniques from 3D point clouds have become widely adopted for discontinuity characterisation of rock masses. These techniques can be broadly categorised into three main types, each with distinct advantages and limitations as summarised in Table 1.

***Table 1.*** *Different discontinuity characterisation techniques.*

| Types | Methods | Advantages | Disadvantages |
|---|---|---|---|
| Clustering | k-Means | Automatic; easy to implement | Flat parametric clustering on point normals; computationally complex indices to estimate the number of sets |
| | Discontinuity Set Extractor (DSE) | Semi-automatic; parameter definition flexibility | Fixed number clustering on density peaks; low inter-cluster separation; overdependence on thresholds |
| | Clustering on Local Point Descriptors (CLPD) | Automatic; high accuracy and robustness | Computationally complex feature set generation; n-dimensional complex clustering; computationally complex indices to estimate the number of sets |
| | Amplitude and Phase Decomposition (APD) | Automatic; high accuracy and efficiency; Feature-based filtering for robustness | Computationally complex indices to estimate the number of sets; Require post-processing filters |
| Segmentation | Region-growing | Automatic | Low efficiency, sensitive to data noise; over-segmentation |
| | Voxelisation | Automatic; Condition-specific high efficiency | Accuracy and efficiency are inversely proportional and dependent on voxel size; requires significant pre-processing and smoothing |
| | Cloud Compare Facets | Efficient fast marching implementation | Over-segmentation; does not cluster identified facets into discontinuity sets |
| Data Driven Learning | Point-based Deep Learning | Automatic; independent of feature generation | Low generalisability; limited training data; overfitting; Unable to identify set variability beyond the trained number of sets; narrow use case |
| | Artificial Neural Network | Automatic; condition-specific accuracy | Computationally complex feature set generation; Unable to identify set variability beyond the trained number of sets; narrow use case |
| | Segment Anything Model | Automatic; independent of feature generation | 2D segmentation model used for 3D segmentation; 2D to 3D transformation only works on specific niche scenarios; unsuitable for real world application |

The first and most widely adopted category is clustering-based approaches, which are among the easiest techniques to implement and, depending on the method, can yield highly accurate and robust results. These techniques generally perform clustering either on point orientations alone or on a combination of orientations and point features to characterise discontinuity sets, commonly employing algorithms such as k-means and k-medoids [18-23]. However, a significant limitation of many such techniques is their reliance on complex validity indices such as the CHI index, Silhouette coefficient, and Xie-Beni index to determine the optimal number of clusters automatically [24]. While effective in principle, this process incurs substantial computational overhead, rendering it impractical for real-world applications involving medium- to large-scale point clouds. Semi-automatic alternatives, such as DSE [25-28], circumvent this by using a predefined cluster count or clusters derived from local maxima in the data. However, this often results in severe over-segmentation and poor inter-cluster separation. Techniques clustering on local point descriptors offer improved robustness by operating in an n-dimensional feature space, but at the cost of significant computational complexity in both feature generation and high-dimensional clustering [29]. On the other hand, techniques like APD introduce unique point features for point filtering to enhance clustering robustness and have demonstrated high accuracy, yet are similarly affected by index complexity [30]. Overall, clustering-based approaches can deliver high accuracy and remain promising for practical structure mapping if their computational costs can be sufficiently reduced.

Segmentation-based techniques represent another widely researched category. In these approaches, the point cloud is first partitioned into discrete planar facets using methods such as region growing [31-33] or voxelisation [34, 35], after which the identified planes are characterised into discontinuity sets. A fundamental limitation of these techniques is their sensitivity to point normal variations, which can cause either over-segmentation in the presence of local noise or under-detection of genuine structural features. Furthermore, the recursive nature of many segmentation algorithms, propagating from seed to neighbouring points, introduces significant computational complexity, limiting their scalability to real-world datasets. Certain implementations, such as those available in CloudCompare Facets, partially address the computational constraint through fast marching estimations, though this often introduces further over-segmentation artefacts [36]. Due to their limited robustness and scalability, segmentation-based techniques are generally not well-suited for practical field implementation.

Finally, a growing body of research has explored data-driven learning approaches for structure mapping. However, these methods are fundamentally constrained by a fatal limitation. Point-based and neural network models are typically trained on data labelled according to a fixed number of discontinuity sets [37-42]. Consequently, they are unable to accommodate variability in the number of sets present, consistently predicting exactly the number of classes seen during training regardless of the actual structural complexity of the test data. While such methods have shown promising results when training, validation, and testing data are drawn from the same point cloud, this experimental design does not reflect real-world conditions, where discontinuity set numbers and configurations vary considerably across sites and geological settings. The absence of open-source or proprietary datasets that are sufficiently diverse and reflect all such set variabilities to train generalisable models further bottlenecks progress in this area. Some approaches attempt to address this limitation by leveraging general-purpose segmentation architectures such as the Segment Anything Model (SAM) to identify planar facets [43]. However, such models are inherently designed for 2D imagery and require 3D-to-2D projection transformations to function on point clouds. This adaptation degrades performance significantly in spatially complex scenes, restricting applicability to clean, simple, single-wall scenarios. As such, data-driven approaches in their current form are not yet suitable for practical real-world structure mapping.

### 2.2. Rock Bolt Identification Techniques

Similar to structure mapping, there has been considerable research done on the automatic identification of rock bolts in 3D point clouds. All such techniques with their detailed advantages and disadvantages have been summarised in Table 2. These techniques strive to solve the problem of the time-consuming and human-biased nature of rock bolt monitoring in the low-light and geometrically complex conditions of underground mine environments.

***Table 2.*** *Different rock bolt identification techniques.*

| Techniques | Advantages | Disadvantages |
|---|---|---|
| Artificial Neural Network | • Rich 58-dimensional manually handcrafted point features<br>• Selective descriptors exploit known bolt cylindrical geometry | • Features are manually engineered, not all significant for bolts<br>• High feature redundancy reduces class separability<br>• High computation complexity<br>• Low robustness to similar geometric structures |
| Course-to-Fine Approach | • Coarse-to-fine strategy reduces false positives<br>• Cluster standardisation improves classification consistency | • Neural network model requires a standard point cluster size as input, making it unsuitable for variable point density<br>• Limited to clean pre-shotcrete scans<br>• Poor generalisation to real-world conditions and environmental variability |
| Multi-Scale Canupo Classifier | • MLS resampling mitigates noise and sparsity effectively<br>• Multi-scale point descriptor captures geometry at multiple dimensions<br>• RANSAC adds geometric verification of bolt candidates | • Moving least squares adds computational overhead<br>• RANSAC sensitive to parameter tuning<br>• Limited cross-environment transferability<br>• Heavily dependent on specific geometry description, and may lead to severe false positives in real-world environment |
| DeepBolt | • Innovative two-stage deep learning architecture explicitly addresses class imbalance and reduces computational complexity<br>• Dynamic graph convolution captures local and global geometric structure<br>• Automatic feature learning<br>• Validated on large-scale real-world mine data | • Requires labelled training data<br>• Generalisation capability across environmental variability in coal mines, although not yet demonstrated in metal mines |
| Cloth Simulation Filter | • Fully geometry-driven pipeline with low computational cost<br>• CSF-based roof-as-ground strategy separates bolt candidates from roof | • Performance dependent on data quality and bolt visibility<br>• Manual roof extraction required, is not fully end-to-end automated<br>• CSF parameters require site-specific tuning<br>• Restricted to approximately planar single-surface processing<br>• Unsuitable for structurally complex or irregular tunnel environments |

The techniques present in the literature can be broadly classified into two categories, namely feature engineering-based techniques and learning-based techniques. Feature engineering–based techniques rely on explicitly designed geometric descriptors to characterise bolt-like structures within point clouds. Early approaches, such as artificial neural networks built on manually crafted feature sets, leverage high-dimensional descriptors and prior assumptions about cylindrical geometry to distinguish bolts from surrounding rock mass [44]. On the other hand, multi-scale methods like CANUPO exploit dimensionality across scales combined with RANSAC-based geometric verification [45]. Filtering approaches such as cloth simulation filtering (CSF) further simplify the problem by separating bolt candidates from the tunnel surface using geometric assumptions, albeit at the cost of full autonomy and site-specific tuning requirements [46]. However, these methods are fundamentally limited by their dependence on hand-crafted features and rigid geometric priors. In real mine environments, where occlusions, shotcrete coverage, irregular surfaces, and noise are prevalent, such assumptions break down. This leads to high feature redundancy, sensitivity to parameter tuning, poor transferability across sites, and ultimately unreliable performance in complex, unstructured conditions. Additionally, several of these techniques incur significant computational overhead due to high-dimensional per-point feature generation, rendering them impractical for large-scale real-world deployment.

In contrast, learning-based approaches, and deep learning methods in particular, aim to overcome these limitations by learning discriminative representations directly from data rather than relying on manual feature definitions. Early learning-based methods, such as coarse-to-fine strategies used for identifying bolts in simple, clean pre-shotcrete civil tunnels, improve classification consistency by progressively refining bolt candidates [47]. However, they are limited by the geometric assumptions and dependence on standardised input requirements, which limit generalisation to real-world environments. One of the most recent deep learning-based methods, termed DeepBolt address these shortcomings by integrating geometric filtering strategies with graph-based segmentation models [48]. The filtering stage directly addresses the severe class imbalance inherent in underground mine point clouds, where bolt points constitute a tiny fraction of the total data, while the graph-based segmentation model adapts to variations in bolt appearance, partial visibility, and environmental noise without relying on strict geometric assumptions.

While the requirement for labelled training data remains a practical constraint, it is a tractable one in the context of rock bolt detection. Unlike discontinuity set characterisation, where training data requirements are complicated by high geological variability in set number and orientation across sites, rock bolts are geometrically consistent objects that are predominantly cylindrical in form with similar physical properties across different mine sites. Furthermore, individual site scans inherently contain large numbers of bolts exhibiting a wide range of angular and spatial configurations, providing rich intra-site geometric diversity within a single training dataset. Crucially, the geometry-sensitive filtering stage substantially reduces background point volume before segmentation, mitigating the impact of environmental variability and simplifying the learning problem to one of accurate detection of a geometrically well-defined cylindrical object.

## 3. Study Area

For this study, point cloud scans were required that capture exposed rock mass with prominent structural features and installed rock bolts for adequate support. However, no open-source dataset currently represents such real-world mine conditions. Therefore, point cloud data of rock mass were collected from tunnels of an Australian hard rock metal mine site using a Leica laser scanner. Figure 1 shows the point cloud scans of three such medium-scale tunnel sections that were acquired. The scans have a nominal point spacing of approximately 0.0025 m and a point density of ~160000 points/m², providing sufficient detail to capture structural features as well as finer elements such as rock bolts. Each tunnel section has a height ranging from 5 to 7 m, a width of 10 to 15 m, and a length between 15 and 25 m. The three scans contain 89, 67, and 84 rock bolts, respectively. The integrated rock support visualisation approach presented in this study is demonstrated on the entirety of Scan 1. The bounding box shown in Figure 1a highlights a region of the tunnel roof wall that is used as a zoomed-in view to present the results in closer detail. This region is also shown in Figure 1b, and is specifically selected due to the abundance of exposed structural features and minimal shotcrete coverage, allowing for clearer visualisation of the results. This section contains 16 bolts, resulting in a higher bolt density compared to other sections of the tunnel, making it a particularly informative region for close-up inspection of the integrated visualisation outputs. Figure 1b also presents a zoomed-in view of the point cloud, illustrating an example of rock bolt points.

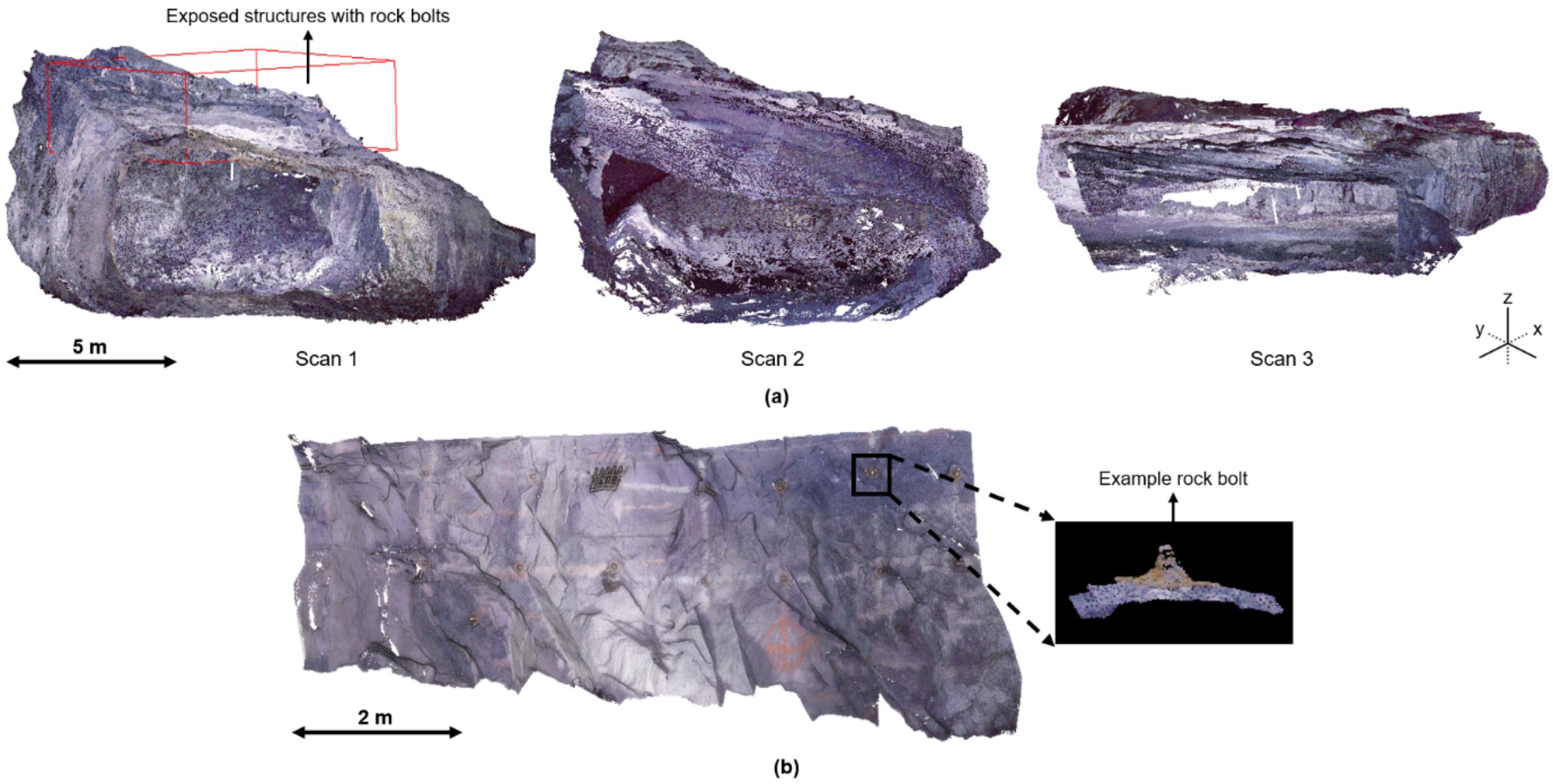

***Figure 1.*** *(a) Point cloud scans of mine tunnel sections. (b) Selected roof wall section with clearly exposed structural features, minimal shotcrete obscuring and visible rock bolts.*

## 4. Methodologies

In this section, a detailed description is presented for the data pre-processing workflow used, the methodology for structure mapping, discontinuity plane fitting, the approach for rock bolt identification, rock bolt orientation estimation, and the integrated visualisation of rock bolt orientations relative to rock mass structural orientations.

### 4.1. Data Pre-Processing

Although the point cloud scans captured by the laser scanner are dense and highly detailed, preprocessing was used to remove any present noise and erroneous outliers. This is critical because rock bolts are small features within the point cloud, and even minor noise can introduce classification errors and degrade final results. Noise may arise from range measurement inaccuracies caused by beam divergence and sensor perturbations. To address this, a statistical outlier removal filter was applied. This filter removes points whose mean distance to their 6 nearest neighbours exceeds one standard deviation from the global average distance, effectively eliminating sparsely distributed outliers. A neighbourhood size of 6 was selected through trial and error, as it provides a minimal yet stable local sample that captures point spacing without over-smoothing fine geometric features. This method was preferred over alternative filtering methods since it is a standard surveying tool with low computational cost and is able to remove spurious points while preserving fine details such as rock bolts and structural features. Additionally, the point cloud was downsampled using a uniform voxel grid approach with a voxel size of 0.02 m. This value was selected empirically, as it substantially reduced file size while preserving fine structural features. In addition to reducing data volume, voxel grid downsampling improves point density uniformity by eliminating redundant points introduced by laser backscattering, and the resulting reduction in file size directly enhances computational efficiency in subsequent processing steps.

Finally, a cloth simulation filter (CSF) was applied to remove floor points from the point cloud, as the tunnel floor contains no structural features or support elements relevant to this study, and its inclusion would only increase computational overhead unnecessarily. The method works by first inverting the point cloud by numerically replacing z-axis coordinates with negative values of themselves and then simulating a flexible cloth, represented as a grid of nodes at a defined spatial resolution, draped over the inverted surface under gravity. Each cloth node is iteratively projected towards the nearest point in the inverted cloud until the cloth conforms to the surface geometry, at which point the simulation reaches equilibrium. Points in the original point cloud that lie within a defined distance of the equilibrium cloth surface are then classified as ground points and removed. The cloth node spacing was empirically set to 50 times the average point spacing, providing a cloth resolution coarse enough to conform to the overall floor geometry without being influenced by localised surface irregularities. The number of iterations was set to 500, consistent with values reported in the literature as sufficient for stable terrain simulation convergence. A classification threshold of 0.5 m was applied, such that points within this distance of the simulated terrain surface are classified as floor and removed, effectively segregating the structural tunnel surfaces of interest from ground-level points.

To exploit local neighbourhood geometry in the processing workflow, a local support region was defined around each point. Following the approach of previous studies [29, 48], the radius of this region was determined as a function of the nominal point spacing PS, as given in Equation 1. This formulation ensures that for point clouds of reasonable density (PS < 0.15 m), the support radius increases monotonically with point spacing, providing an optimal neighbourhood size for accurate local geometric representation. The nominal point spacing of the downsampled point cloud was estimated using 3D Delaunay triangulation of the point cloud, yielding a *PS* of 0.012 m and a corresponding support radius of 0.058 m.

$$Spherical\ Support\ Radius = PS \times (5 - 16 \times PS) \tag{1}$$

Eigenvalue descriptors (EVD) are necessary to characterise local surface geometry in the point cloud based on the spatial distribution of points within a neighbourhood. Eigenvalue descriptors are preferred over other point descriptors due to their direct geometric interpretability, significantly lower computational complexity and high robustness to data noise [49, 50]. While few other descriptors may provide a richer surface description, in the case of underground mine point clouds, geometric simplicity and computational efficiency empirically make EVD the better choice as the initial step for both structure mapping and rock bolt identification. For each point in the point cloud, three eigenvalues ($\lambda_1 \geq \lambda_2 \geq \lambda_3$) were computed by applying Principal Component Analysis (PCA) to the local neighbourhood defined by the support radius in Equation 1. PCA constructs the 3×3 covariance matrix of the neighbourhood point distribution and performs eigen-decomposition to extract three orthogonal principal axes and their associated variances. The resulting eigenvalues represent the spread of points along each principal axis and serve as the basis for computing the eigenvalue descriptors for surface geometry representation in the processing workflow. Additionally, the eigenvector corresponding to the smallest eigenvalue $\lambda_3$, representing the direction of least variance within the local neighbourhood, is extracted as the unit normal vector N for each point, as points on a surface vary least in the direction perpendicular to it. These per-point normals are subsequently utilised in the processing workflow to identify orientations of the structures.

### 4.2. Overview of the Processing Workflow

A general overview of the complete processing workflow is presented in Figure 2. The process starts by feeding the pre-processed input point cloud with its calculated per-point eigenvalues and point normal simultaneously into two complementary processing pipelines, namely the structure mapping and rock bolt identification workflows. In the structure mapping pipeline, discontinuity sets of coherent orientations, geometry, and mechanical properties are automatically characterised, from which the dominant discontinuity plane orientations are extracted. A best-fit 3D plane is then computed for each identified discontinuity set, providing a geometric representation of the rock mass structure in 3D space. In the rock bolt identification pipeline, rock bolts are semantically segmented within the point cloud to identify and localise all installed bolts to determine their spatial positions. The orientation of each segmented bolt is subsequently estimated and represented as an orientational vector in 3D space. The outputs of both workflows are subsequently integrated to produce the integrated rock support visualisation. In the first visualisation, bolt orientation vectors are overlaid directly onto their discontinuity planes in 3D space, providing a spatially explicit overview of support structure geometry relative to rock mass structure, facilitating a critical 3D visual model for geotechnical assessment of support design geometric effectiveness. In the second visualisation, bolt vector orientations and discontinuity poles are jointly projected onto a 2D stereonet, enabling direct comparison of bolt angles against the prevailing structural discontinuity poles, thereby forming a visualisation graph for identification of misaligned or ineffective bolts. Each component of the workflow is described in detail in Sections 4.3 – 4.7.

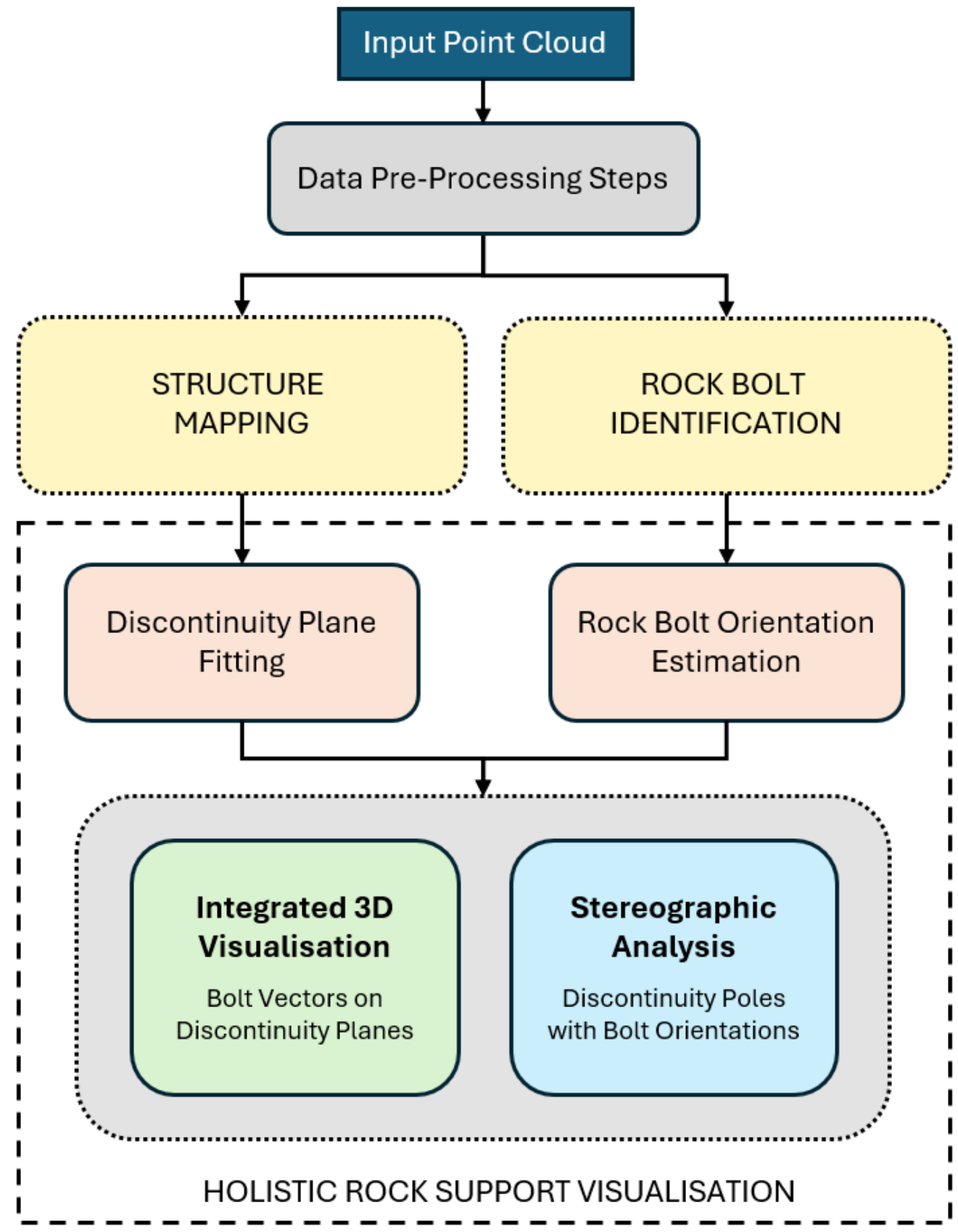

***Figure 2.*** *General overview of the processing workflow.*

### 4.3. Structure Mapping

In this study, a clustering-based structure mapping approach is adopted, as the literature demonstrates that clustering-based techniques offer the most favourable balance between accuracy and robustness for real-world applications. The algorithmic architecture of the structure mapping technique used, leading to the discontinuity plane fitting for the integrated rock support visualisation, is presented in Figure 3. The technique has been tailored to minimise the computational overhead inherent to standard clustering-based methods, making it practical for real-world applications such as the one presented in this study. Drawing on the learnings and the limitations identified in the literature as described in Section 2.1, two key design decisions were made. First, a point filtering strategy is incorporated prior to orientation clustering to improve robustness to point orientations, ensuring that only geometrically relevant points contribute to discontinuity set characterisation. Second, a clustering algorithm is employed that does not rely on complex validity indices for automatic determination of the number of discontinuity sets, avoiding the computational bottleneck that renders many existing techniques impractical at scale.

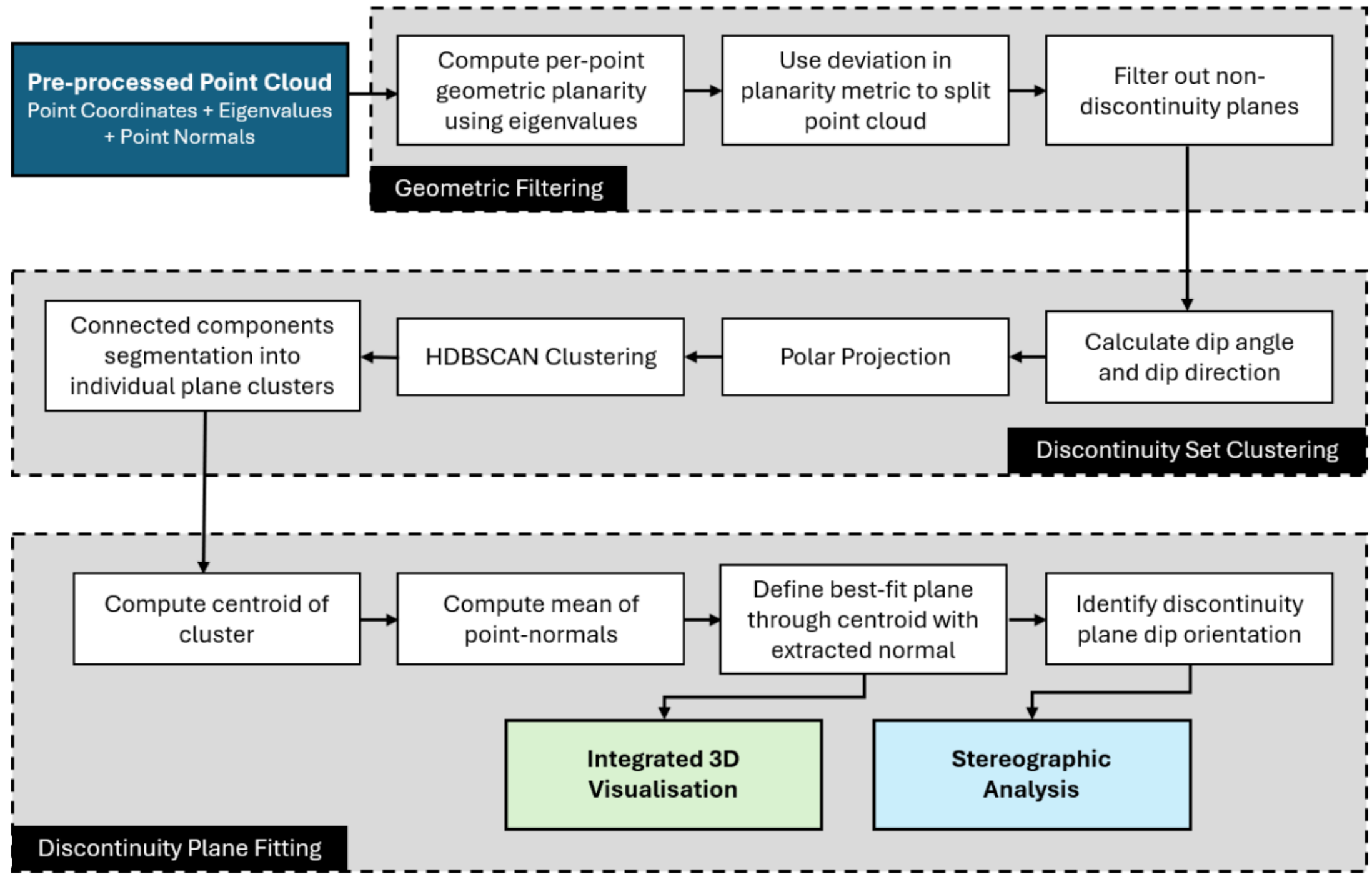

***Figure 3.*** *Algorithmic architecture for structure mapping and discontinuity plane fitting.*

For point filtering, a geometric filtering strategy was employed to retain only points belonging to discontinuity planar surfaces while removing non-planar points and noise, thereby improving the robustness of subsequent orientation clustering. Various point filtering strategies have been explored in the literature, employing different point metrics for this purpose, for instance, the APD method filters points based on the signal characteristics of a point's neighbourhood in the frequency spectrum to isolate planar regions. The common underlying principle across such approaches is to separate geometrically planar points from irrelevant non-planar points prior to clustering. For this study, an eigenvalue descriptor-based geometric filtering strategy was selected for two complementary reasons. First, as established in Section 4.1, eigenvalue descriptors offer direct geometric interpretability, computational efficiency, and robustness to noise, properties that make them well-suited for this filtering task. Second, since eigenvalues were already computed during preprocessing and are simultaneously required for rock bolt identification in the parallel pipeline, reusing them for point filtering avoids redundant feature computation and eliminates additional computational overhead, a key design consideration for practical scalability. Using the per-point eigenvalues computed in Section 4.1, the planarity descriptor was calculated for each point using Equation 2. Planarity quantifies how well the local neighbourhood of a point resembles a plane, ranging from 0 to 1, with 1 representing a perfect geometric plane and 0 representing a geometric line, sphere, or isotropic noise. This geometric property was leveraged to discriminate discontinuity planar surfaces from non-planar features such as discontinuity edges, rock bolt protrusions, and high-curvature artefacts. Through empirical observation across multiple diverse datasets, points exhibiting planarity values exceeding 0.8 were found to correspond to discontinuity planar regions. This value was therefore adopted as the filtering threshold, retaining only high-planarity points and discarding all others. The resulting filtered point cloud contains predominantly discontinuity planar points, significantly reducing noise and irrelevant geometric features and improving the robustness of the subsequent discontinuity set clustering step.

$$Planarity = (\lambda_2 - \lambda_3) / \lambda_1 \quad (2)$$

Following geometric filtering, discontinuity set clustering was performed to characterise the filtered point cloud into discrete discontinuity sets, from which representative discontinuity planes are subsequently derived. To enable orientation-based clustering, the per-point normals N of the filtered points were first converted into geological orientation descriptors, namely dip angle and dip direction, using Equation 3. This representation is a standard geological convention for describing orientations, and when plotted on a stereonet, they accurately represent 3D geometry in a 2D plane without loss of spatial context. Dip angle defines the maximum angle of inclination of a planar surface measured downward from the horizontal, while dip direction defines the compass azimuth in which the plane descends downwards. However, these orientation descriptors are inherently cyclic in nature, where a dip direction of 0° and 360° are geometrically identical, yet numerically distant. If clustering is performed directly on raw angular values, this boundary discontinuity introduces artificial separation between orientations that are in reality equivalent, leading to spurious clusters at the angular boundary. To address this, a polar transformation was applied to the dip angle and dip direction values using Equation 4, mapping the cyclic angular data into Cartesian space. This transformation preserves the true circular continuity of the orientations, ensuring that clustering operates on a geometrically consistent representation and eliminating boundary artefacts that would otherwise degrade set characterisation accuracy.

$$Point\ Orientation\ Pair\ [DA\ DD] = \begin{cases} \begin{cases} DA = cos^{-1}(N_z) \\ DD = tan^{-1}\left(\frac{N_x}{N_y}\right) \end{cases}, for\ DA \le 90° \\ \begin{cases} DA = 180° - DA \\ DD = DD + 180° \end{cases}, for\ DA > 90° \end{cases}$$

*Where $N_x$, $N_y$ and $N_z$ are the vector components of the normal* (3)

$$radius\ R = \frac{\sin\ (DA)}{1 + \cos\ (DA)}$$

$$D_{Px} = R \times \sin\ (DD)$$

$$D_{Py} = R \times \cos\ (DD)$$

*Where [ $D_{Px}$ $D_{Py}$ ] is the transformed orientation pair* (4)

For the clustering step, a method was required that could automatically identify discontinuity sets in the transformed orientation data without relying on computationally expensive validity indices for optimal cluster number determination. For this reason, HDBSCAN (Hierarchical Density-Based Spatial Clustering of Applications with Noise) was selected as the clustering technique. HDBSCAN is an unsupervised, non-parametric, hierarchical density-based clustering algorithm that displays high accuracy for datasets with varying density distributions, a characteristic inherent to point orientation data [51, 52]. As a hierarchical technique, HDBSCAN determines the number of clusters automatically without requiring any external index, directly addressing the computational bottleneck identified in standard clustering-based approaches. Furthermore, its density-based nature enables it to explicitly model noise as unclustered points rather than assigning all points to a cluster as flat techniques such as k-means do, eliminating the need for separate majority filtering steps for outlier removal. HDBSCAN requires only two parameters. The first is the minimum cluster size, which defines the minimum number of points required for a group to be considered a valid cluster and was set to 10000 to ensure that geologically insignificant stray orientation groups are not falsely identified as discontinuity sets. The second is the minimum sample size, which defines the local density threshold for identifying core points and was empirically set to 100, providing an optimal balance between being too restrictive, which would miss valid low-density discontinuity sets, and too permissive, which would incorrectly absorb noise into clusters. Notably, unlike other density-based techniques such as DBSCAN, HDBSCAN does not require specification of a neighbourhood distance threshold (ε), a parameter that is particularly difficult to tune in orientation data with highly varying densities, instead circumventing this requirement through its hierarchical dendrogram tree structure.

Once the filtered points were characterised into discontinuity sets, a secondary segmentation step was performed within each set to isolate individual discontinuity planes, as multiple spatially separated planes may share the same orientation set. Since distinct planes within a set are separated by physical distance, a Euclidean distance-based connected component segmentation was applied within each set to delineate individual planes. Connected component segmentation groups spatially connected points using an octree spatial partitioning structure, with the octree cell size set to the radius of influence determined, ensuring that the spatial connectivity criterion is consistent with the local neighbourhood scale used throughout the processing workflow. Following segmentation, only planes constituted of more than 100 points were retained, as smaller segments lack sufficient geometric extent to carry meaningful geological significance and are treated as residual noise.

Following discontinuity set characterisation, the discontinuity plane fitting is performed in 3D space to obtain an explicit geometric representation of each plane, which is subsequently utilised in the integrated 3D and stereographic visualisation framework. The plane fitting technique and its integration with the rock bolt orientation vectors within the integrated visualisation framework are discussed in detail in Section 4.5.

#### 4.4. Discontinuity Plane Fitting

Once individual discontinuity planes have been isolated within each characterised discontinuity set as described in Section 4.3, a best-fit plane is computed for every identified plane across all discontinuity sets. An overview of this discontinuity plane fitting process, following the discontinuity set characterisation, can also be seen in Figure 3. The fitting procedure comprises four sequential steps as described below.

a) Step 1 – Compute cluster centroid

For each identified discontinuity plane cluster k, the centroid $\mu_k$ is calculated using Equation 5, representing the geometric centre of the cluster in 3D space and serving as the anchor point for the fitted plane.

$$\mu_k = \frac{1}{N_k}\sum_{i=1}^{i=N_k} q_i$$

*Where* $q_i = (x_i, y_i, z_i)^T$ *are the* $N_k$ *points belonging to point cluster k.* (5)

b) Step 2 – Compute mean of point normals

The plane normal for each identified discontinuity plane k is estimated as the mean of the unit normals of all points within the cluster, computed using Equation 6. This approach is preferred over recomputing a cluster-level covariance matrix, as it avoids redundant computation given that point normals were already determined during preprocessing. Furthermore, since all points belonging to a single discontinuity plane are predominantly oriented in the same direction, the mean normal provides an accurate and geometrically consistent estimate of the plane orientation.

$$\bar{v} = \frac{1}{N_k}\sum_{i=1}^{i=N_k} v_i\,, \qquad \hat{v} = \frac{\bar{v}}{\|\bar{v}\|}$$

*where* $\boldsymbol{v}_i$ *is the oriented unit normal of point* $i$ *retrieved from preprocessing (eigenvector of* $\lambda_3$*), and* $\hat{\boldsymbol{v}}$ *is the unit plane normal of the fitted plane.* (6)

c) Step 3 – Define best-fit plane

Using the computed centroid and normal, a best-fit 3D plane representing the discontinuity surface is defined in 3D space using Equation 7. These fitted planes serve as 3D geometric representations of the structural features present in the rock mass, providing the primary visual elements for the integrated rock support visualisation. For visualisation purposes, the spatial extent of each plotted plane is bounded by the physical dimensions of its corresponding point cluster.

$$\hat{v} \cdot (p - \mu_k) = 0$$

*Where* $p$ *is any point lying on the plane.* (7)

d) Step 4 – Compute plane orientation

Once the best-fit plane is defined in 3D space, its dip angle and dip direction are computed for stereonet plotting using the vector components of $\hat{v}$ in Equation 3. The discontinuity set membership of each plane is preserved throughout, enabling colour-coded representation of planes by set affiliation in both the 3D visualisation and the stereonet, facilitating direct visual comparison of structural orientations.

### 4.5. Rock Bolt Identification

Given the complex and variable nature of rock bolt protrusions in real-world point clouds, where bolts are tiny, partially occluded by shotcrete, and embedded within geometrically cluttered mine environments, a deep learning-based approach is the most appropriate choice for this study as suggested by the literature in Section 2.2. Specifically, an adapted version of DeepBolt [48] is used as the rock bolt identification technique, as its two-stage architecture is well-suited to handling class imbalance, environmental variability, and partial bolt visibility. Furthermore, to improve generalisation beyond the coal mine environments on which DeepBolt was originally trained, the model is retrained in this study with additional data from metal mine point clouds, broadening its applicability to the specific scanning conditions and geological settings encountered in this work. A detailed overview of the rock bolt identification workflow utilising DeepBolt, followed by the rock bolt orientation estimation, is presented in Figure 4.

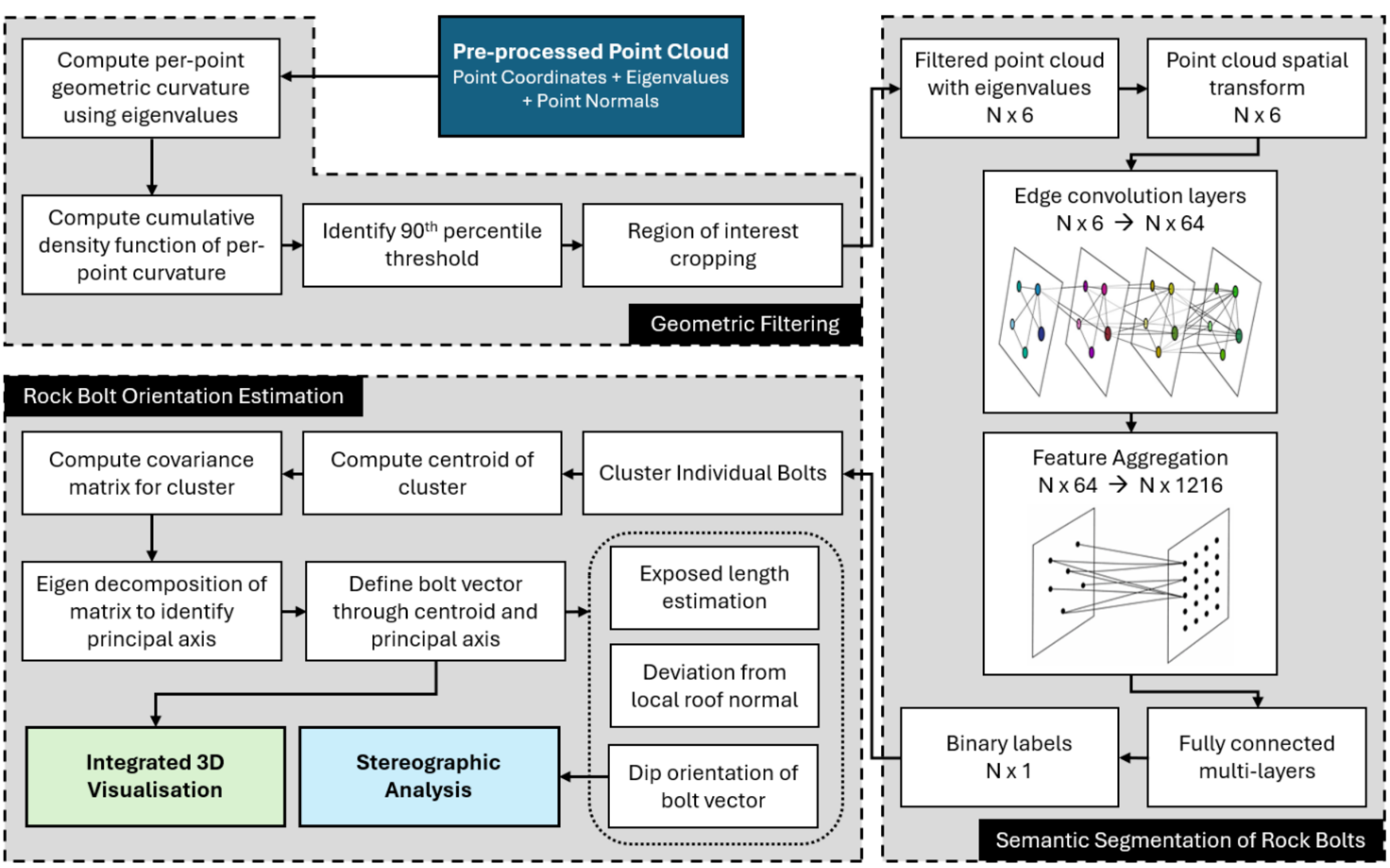

***Figure 4.*** *Algorithmic architecture for rock bolt identification and orientation estimation.*

The rock bolt identification architecture employs a two-stage deep learning pipeline. Similar to the discontinuity characterisation technique described in Section 4.3, the first stage is a geometric filtering step that leverages the per-point eigenvalues computed in Section 4.1. Geometric curvature is calculated for each point using Equation 8, serving as the primary discriminative descriptor for this stage. Curvature quantifies the degree of local surface flexion within a local neighbourhood, hence bolt protrusions, being cylindrical in nature, naturally exhibit high curvature values relative to the predominantly planar background surfaces such as rock faces and shotcrete, making it well-suited for separating bolt candidates from background points. To determine the filtering threshold, the cumulative density function (CDF) of per-point curvature values is computed across the point cloud. Empirical observation across datasets from multiple mine sites suggested that bolt protrusions consistently reside within the top 8th percentile of curvature values. Accordingly, the 90th percentile is adopted as a conservative threshold providing a safe buffer margin for complete bolt preservation. Points exceeding this threshold are retained as high-curvature candidates, while the remainder are discarded. The retained candidates are then spatially refined through connected component segmentation using the same octree settings as Section 4.3. Clusters below 100 points are discarded as stray artefacts, clusters above 400 points are passed through directly as they represent localised high-curvature regions too large to be individual bolt protrusions but likely to contain bolts within them, and for all remaining clusters a spherical region of interest of radius 0.15 m is extracted around the cluster centroid, recovering any bolt points inadvertently removed during thresholding given that visible protrusion lengths ranged between 0.05 and 0.25 m in the dataset. The primary objective of this filtering stage is twofold: a) to substantially reduce background point volume passed to the segmentation model for computational efficiency, and b) to mitigate the severe class imbalance between bolt and non-bolt points, simplifying the subsequent learning task.

$$Curvature = \lambda_3 \,/\, (\lambda_1 + \lambda_2 + \lambda_3) \quad (8)$$

The filtered point cloud, containing significantly reduced background points, is passed to the second stage, semantic segmentation of rock bolts using a dynamic graph convolution neural network. The input to the segmentation model is an N×6 matrix per point, comprising the three spatial coordinates (X, Y, Z) and the three eigenvalues ($\lambda_1$, $\lambda_2$, $\lambda_3$) computed during preprocessing. Incorporating eigenvalues alongside point coordinates enriches the geometric representation of the input, accelerating model convergence by providing explicit local shape context. The input first passes through a point cloud spatial transform block, which estimates and applies a 3×3 transformation matrix to align the point set to a canonical space, ensuring invariance to rotations and translations. The transformed features are then processed through a series of edge convolution layers, where a dynamic k-nearest neighbour graph is constructed in feature space at each layer and edge features between each point and its neighbours are computed and processed through a multi-layer perceptron (MLP). By dynamically reconstructing the local neighbourhood graph at each layer in the evolving feature space rather than using fixed spatial neighbourhoods, the model adapts to variations in local geometry and point density, improving feature discrimination between bolt and background points. The features extracted across the three edge convolution layers, capturing local detail, mid-range structure, and global context respectively, are concatenated through a feature aggregation step, yielding a unified feature vector that preserves both fine and coarse geometric information. Finally, the aggregated features are passed through fully connected multi-layer perceptron layers, producing a single per-point logit that is activated through a weighted sigmoid binary cross-entropy loss function to yield a bolt probability score. Points with a probability exceeding 0.5 are assigned a binary bolt label, while the remainder are classified as non-bolt, completing the semantic segmentation of rock bolts in the point cloud.

The segmentation model used in DeepBolt was originally trained on labelled geometrically filtered point clouds from coal mine environments, comprising 1764 labelled bolt protrusions. The model demonstrated strong generalisation within coal mining conditions. However, while bolt protrusions exhibit broadly similar geometric characteristics across coal and metal mining environments, the residual background structures that pass through the filtering stage can differ between the two settings. To improve generalisation to the metal mine data used in this study, the model was retrained using an augmented dataset combining the original coal mine training data with manually annotated geometrically filtered point clouds from Scans 2 and 3 of the metal mine datasets, bringing the total labelled bolt count to 1915. The annotated metal mine data, apart from incorporating additional context on bolt geometry and spatial distribution characteristic of metal mining conditions, also captures variability in surface roughness, shotcrete coverage, and background structural complexity typical of hard rock environments within the residual background points. All 89 bolts in Scan 1 were reserved exclusively for testing and withheld from training entirely to prevent overfitting and provide an unbiased evaluation of model performance. Only the bolt protrusions have been considered in the study, and the face plates have been excluded since they are often obscured by shotcrete coverage. Training was conducted using 10-fold cross-validation to monitor convergence, with the model stabilising at 32 epochs, which is consistent with the original DeepBolt study. The training and validation loss curves exhibited parallel trends throughout, with no overlaps after convergence, confirming stable convergence without overfitting, proving the model's ability to accurately predict bolts in unseen data. This retraining step extends the applicability of the model to the scanning conditions and geological settings of metal mine environments considered in this study.

Using the trained semantic segmentation model, rock bolt protrusions are identified and localised within the point cloud. The identified bolt points are subsequently passed to the rock bolt orientation estimation step, which forms an integral component of the integrated rock support visualisation framework and is described in detail in Section 4.6.

### 4.6. Rock Bolt Orientation Estimation

After the rock bolt points have been identified through semantic segmentation as described in Section 4.5, rock bolt orientation estimation is performed so that the rock bolts can be represented as representative vectors in 3D space for integrated visualisation and stereographic analysis. An overview of this procedure is also shown in Figure 4 following the rock bolt identification stage. The orientation estimation comprises the sequential steps described below.

a) Step 1 – Cluster individual bolts

The semantic segmentation model identifies all points corresponding to rock bolts within the filtered point cloud. To isolate individual bolt components, connected component segmentation is applied to the identified bolt points using settings same as those described in Section 4.3. This step separates the detected bolt points into individual bolt clusters for subsequent orientation estimation.

b) Step 2 – Compute cluster centroid

Similar to step 1 of discontinuity plane fitting, the centroid $\mu_b$ is calculated for each identified rock bolt protrusion cluster b, consisting of $N_b$ points, using Equation 5. This centroid serves as the anchor point for defining the bolt vector and is also used for estimating the protrusion length of the candidate bolts.

c) Step 3 – Compute covariance matrix

The 3×3 covariance matrix $\Sigma_b$ of each bolt cluster is then computed to capture the spatial variance of the cluster points about the centroid along the three principal directions. This is conceptually similar to the principal component analysis applied earlier in the workflow; however, in this case, the covariance matrix is computed using the entire protrusion cluster rather than a local neighbourhood. This is necessary because the objective here is to identify the dominant elongation direction of the bolt, as bolt protrusions form elongated cylindrical features that extend approximately along the bolt insertion direction. The covariance matrix, therefore captures the overall geometric distribution of the protrusion and provides the basis for extracting its dominant orientation in the subsequent eigen-decomposition step. This approach is preferred over a simple best-fit line, as it estimates the overall elongation direction of the cluster rather than forcing a line through the points. As a result, it is more robust in cases where the protrusion is partially occluded or where the point distribution is uneven or asymmetric.

$$\Sigma_b = \frac{1}{N_b - 1}\sum_{i=1}^{N_b}(q_i - \mu_b)(q_i - \mu_b)^T \tag{9}$$

d) Step 4 – Eigen-decomposition to identify principal axis

Eigen-decomposition is performed on $\Sigma_b$ to extract the three principal axes and their associated eigenvalues, as given in Equation 10. Since bolt protrusions are cylindrical elongated structures, the variance along the bolt axis is significantly larger than in the two perpendicular directions, resulting in a dominant largest eigenvalue $\eta_1 \gg \eta_2 \approx \eta_3$. The eigenvector $\mathrm{u}_1$corresponding to $\eta_1$ therefore defines the principal axis, the dominant elongation direction of the bolt protrusion in 3D space.

$$\Sigma_b\, \mathrm{u}_l = \eta_l \mathrm{u}_l, \qquad l = 1,2,3\ (\eta_1 \geq \eta_2 \geq \eta_3)$$

*Where* $\eta_1,\ \eta_2,\ \eta_3$ *are the ordered eigenvalues and* $u_1, u_2, u_3$ *are the corresponding orthonormal eigenvectors.* (10)

e) Step 5 – Define bolt vector through centroid and principal axis

The orientation of each bolt is represented as a vector $\mathrm{r}_b$ passing through the cluster centroid $\mu_b$in the direction of the principal axis $\mathrm{u}_1$, as given in Equation 11. This provides a precise and compact 3D representation of each bolt's spatial orientation and position, serving as the primary geometric descriptor for the integrated visualisation and stereonet analysis. For 3D visualisation, the vector length is limited to 2 m and rendered as a cylindrical solid with a diameter of 20 mm, thereby defining the bolt representation in 3D space.

$$\mathrm{r}_b = \mu_b + t \cdot \mathrm{u}_1,\ t \in \mathbb{R} \tag{11}$$

f) Step 6 – Exposed length, deviation from local roof normal, and dip orientation estimation

The exposed length of each bolt protrusion is estimated by projecting all cluster points onto the principal axis $\mathrm{u}_1$and computing the extent between the minimum and maximum projections, as given in Equation 12. This yields a direct geometric estimate of the visible protrusion length of the bolt, which is a key quality assessment parameter indicating whether the bolt has been installed to the correct depth.

$$s_i = (q_i - \mu_b) \cdot \mathrm{u}_1$$

$$L_b = \max(s_i) - \min(s_i)$$

*Where* $s_i$ *is the scalar projection of each point onto the principal axis and* $L_b$ *is the estimated exposed length of bolt* $b$*.* (12)

The angular deviation of each bolt from the local roof surface normal is computed using Equation 13. This deviation angle $\theta_b$ quantifies how far the installed bolt departs from the ideal perpendicular installation direction relative to the rock surface, serving as a direct and quantitative indicator of installation quality. Bolts with large deviation angles may provide reduced anchoring effectiveness and can be flagged for inspection.

$$\theta_b = \arccos(| \mathrm{u}_1 \cdot \hat{\mathrm{v}}_{\mathrm{roof}} |)$$

*where* $\hat{v}_{roof}$ *is the unit normal of the local roof surface at the bolt location.* (13)

Finally, the dip angle and dip direction of each bolt orientation vector are computed using the vector components of the bolt protrusion axis $\mathrm{u}_1$ in Equations 3, expressing the bolt orientation in the same geological coordinate system as the discontinuity planes. This ensures that bolt orientations and discontinuity plane orientations are directly comparable when jointly plotted on the stereonet in the integrated visualisation framework, enabling geotechnical assessment of the angular relationship between installed bolts and the prevailing structural fabric of the rock mass.

### 4.7. Integrated 3D Visualisation and Stereographic Analysis

Following the estimation of discontinuity planes and rock bolt orientation vectors, the outputs of both processing pipelines are integrated to form the final stage of integrated rock support visualisation. This integration is realised through two complementary representations, namely a spatially explicit integrated 3D visualisation and a stereographic projection-based analysis.

In the integrated 3D visualisation, the fitted discontinuity planes are rendered as bounded planar surfaces in 3D space, colour-coded by discontinuity set membership, while the rock bolts are represented as cylindrical vectors defined by their centroids and principal orientation axes. This direct spatial overlay enables intuitive assessment of the geometric relationship between installed bolts and the surrounding structural fabric of the rock mass. By preserving the true spatial configuration of both structural features and support elements as captured in the point cloud, the visualisation facilitates immediate identification of key geometric geotechnical conditions, including bolts intersecting active discontinuity planes, sub-optimal bolt installations relative to structural planes, and areas where prominent structural features remain unsupported. This enables engineers and mine operators to assess, in 3D space, which structural features are adequately supported and which remain exposed and potentially prone to displacement or rockfall, which can provide an actionable basis for support design review and maintenance planning. In addition to geometric representation, quantitative bolt-level metrics, such as exposed protrusion length and angular deviation from the local roof normal, can also be integrated into the visualisation. These attributes are encoded through colour mapping of the bolt representations, enabling simultaneous spatial and quality-based assessment of the installed support system and providing further insight into bolting effectiveness.

A discrete fracture network (DFN) like visualisation was avoided for the 3D representation because it describes rock mass structure within an idealised representative volume, typically a prismatic or cylindrical domain, and requires fracture-network parameter assumptions such as trace length, intensity, and persistence, or  require supporting in-situ borehole data. Even when DFNs are derived solely from laser scans, these parameters must still be inferred from surface observations, meaning the resulting model is a statistical or deterministic abstraction rather than an exact representation of the observed excavation surface. DFN models are valuable for volumetric fracture-network analysis, but they do not directly represent the actual exposed excavation surface assessed by a geotechnical engineer and are therefore less suitable for the integrated evaluation of surface support interaction presented in this study. In contrast, the proposed framework operates entirely on measured laser-scan geometry and visualises only what is physically observable at the excavation boundary, providing an automated, low-cost, surface-faithful assessment of structural features and installed bolts without subsurface assumptions or cumbersome in-situ data acquisition, while reducing manual interpretation effort, operator subjectivity, and the need for prolonged human assessment of support structures in low-light or hazardous underground conditions.

In addition to the 3D representation, a stereographic analysis is performed to enable quantitative orientation-based comparison of bolt vectors relative to discontinuity sets. The dip orientations of the fitted discontinuity planes and the identified rock bolt orientation vectors are projected jointly onto a lower-hemisphere equal-angle stereonet using a consistent geological coordinate system (dip angle and dip direction), as computed in Sections 4.4 and 4.6 respectively. Discontinuity planes are represented as poles colour-coded by set membership, while bolt orientation vectors are overlaid as a distinct symbol set. This unified projection enables direct evaluation of the angular relationship between individual bolt orientations and each major discontinuity set, facilitating identification of bolts that are optimally oriented to intercept and restrain potential failure planes, as well as those deviating significantly from the expected installation direction, which may indicate installation defects or compromised anchoring performance. The stereonet representation also allows rapid visual assessment of the overall consistency of bolt installation across a tunnel section, and can be used to identify systematic angular biases in bolting patterns relative to the prevailing structural fabric.

The combination of spatially explicit 3D visualisation and stereographic projection provides a comprehensive and complementary framework for assessing rock support systems. While the 3D visualisation offers direct geometric and spatial understanding in physical space, enabling assessment of which structures are supported and which remain exposed, the stereonet enables rigorous orientation-based analysis in a compact 2D representation suitable for standard geotechnical reporting. Together, these two views constitute the integrated rock support visualisation proposed in this study, enabling both qualitative and quantitative evaluation of bolt geometric effectiveness relative to the structural fabric of the rock mass directly from point cloud data.

## 5. Results and Discussions

### 5.1. Structure Mapping

The structure mapping results for Scan 1 are presented in Figure 5. Figure 5a shows the discontinuity sets characterised across the full scan, with six distinct discontinuity sets identified and a total of 196 individual discontinuity planes delineated across all sets. Each discontinuity set is visualised using a unique colour as indicated in the legend, which also reports the number of individual planes identified within each set. For closer inspection, the structure mapping results are additionally visualised on the zoomed-in roof wall section in Figure 5b, where individual discontinuity plane clusters are rendered in unique colours to clearly delineate the planes present in the data.

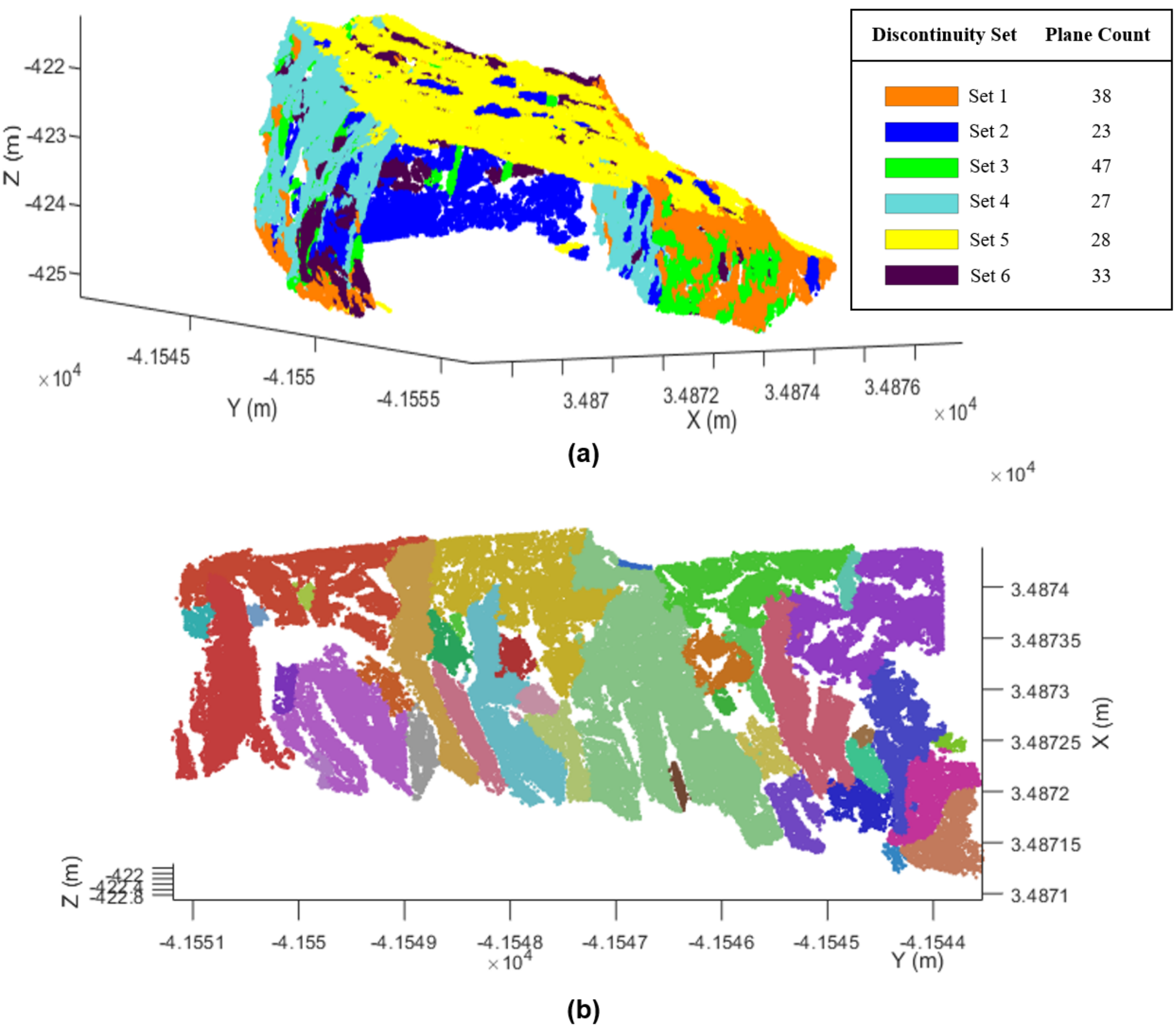

***Figure 5.*** *Structure mapping results on Scan 1 point cloud. (a) Discontinuity sets characterised using*

*the utilised approach and coloured based on set legend. (b) Identified individual discontinuity plane clusters shown in the zoomed-in roof wall section.*

The visual results demonstrate the robustness of the adopted approach. High-curvature artefacts are effectively eliminated from the characterisation output, and individual discontinuity planes exhibit clear spatial separation with minimal inter-cluster overlap, indicating accurate plane delineation. This robustness arises from the combined effect of the three processing stages. The geometric filtering removes high-curvature artefacts and steep discontinuity edges prior to clustering, ensuring that only geometrically planar points contribute to the orientation analysis and creating clean spatial boundaries between adjacent planes. The HDBSCAN clustering then operates on the filtered orientation data, identifying only statistically significant orientation clusters with low angular dispersion about their poles while treating noisy or sparsely distributed orientations, arising from residual sensor noise or insignificant outlier points, as unclustered noise, preventing them from being falsely assigned to discontinuity sets. Finally, the connected component segmentation removes any residual clusters comprising fewer than 100 points, discarding geologically insignificant fragments and isolated noise points that may have encroached into otherwise well-defined planes. The combined effect of these three stages produces the clearly separated, accurate, and geologically meaningful discontinuity characterisation results observed in Figure 5.

To evaluate the accuracy of the discontinuity characterisation, the automated results were compared against a manually derived reference dataset. Direct field measurements were not feasible due to the scale and complexity of the Scan 1 study area; therefore, the CloudCompare virtual compass tool was used to generate the validation dataset. This tool emulates a field compass–clinometer by allowing manual selection of planar facets and computing their dip angle and dip direction, and is widely used in mine surveying as a digital alternative to physical measurements. Within each identified discontinuity set, 10 to 15 representative planes were manually selected using the virtual compass tool, resulting in a total of 55 planes. The orientations obtained from these manually fitted planes were compared with those derived from the automated approach, and the absolute errors for each set are summarised in Table 5. The overall error was found to be 2.33° for dip angle and 3° for dip direction, which falls within the expected range of subjective variability associated with manual measurements. These results indicate that the proposed method achieves a satisfactory level of accuracy comparable to manual interpretation, demonstrating its reliability for discontinuity characterisation and its suitability for subsequent integration within the integrated visualisation framework.

***Table 3.*** *Characterised discontinuity set average orientation validation.*

| Discontinuity Set | Number of Discontinuity Planes | Average orientation (dip angle/dip direction) (°) | | |
|---|---|---|---|---|
| | | Utilised Approach (Automatic) | CloudCompare Virtual Compass (Manual) | Absolute Error |
| 1 | 38 | 68 / 114 | 70 / 120 | 2 / 6 |
| 2 | 23 | 75 / 235 | 72 / 237 | 3 / 2 |
| 3 | 47 | 70 / 315 | 70 / 311 | 1 / 4 |
| 4 | 27 | 62 / 276 | 60 / 278 | 2 / 2 |
| 5 | 28 | 35 / 112 | 39 / 111 | 4 / 1 |
| 6 | 33 | 66 / 58 | 66 / 61 | 2 / 3 |

### 5.2. Rock Bolt Identification

Figure 6 presents the visual results of rock bolt identification, where the semantically segmented bolt points and their corresponding locations are overlaid on the point cloud. The identified bolts are distributed across the excavation surface and exhibit varying exposed protrusion lengths, indicating consistent detection performance throughout the scan with minimal noise in the segmentation output. The zoomed-in inset highlights an example of an identified bolt, showing that the method successfully captures the clustered point signature of the protrusion while avoiding false identification of surrounding background points. The preserved cylindrical geometry of the protrusion further demonstrates the quality of the segmentation. The results were obtained for Scan 1, which was withheld entirely from model training and used exclusively for testing. Since the semantic segmentation model was trained using filtered scans from both coal and metal mine environments, but did not include Scan 1, the results indicate that the model generalises well to unseen data rather than overfitting to the training set. The method is therefore able to accurately identify and segment bolt points from surrounding structures under realistic mining conditions, highlighting its practical applicability across varying underground environments and its effectiveness in distinguishing bolt objects from background noise and non-bolt features. The absence of bolts near the farthest wall region of the point cloud is consistent with the excavation state of the tunnel, where that section remains newly exposed and has not yet undergone support installation. As excavation progresses, bolting would be expected to extend into these regions, explaining the lack of visible bolts in that area.

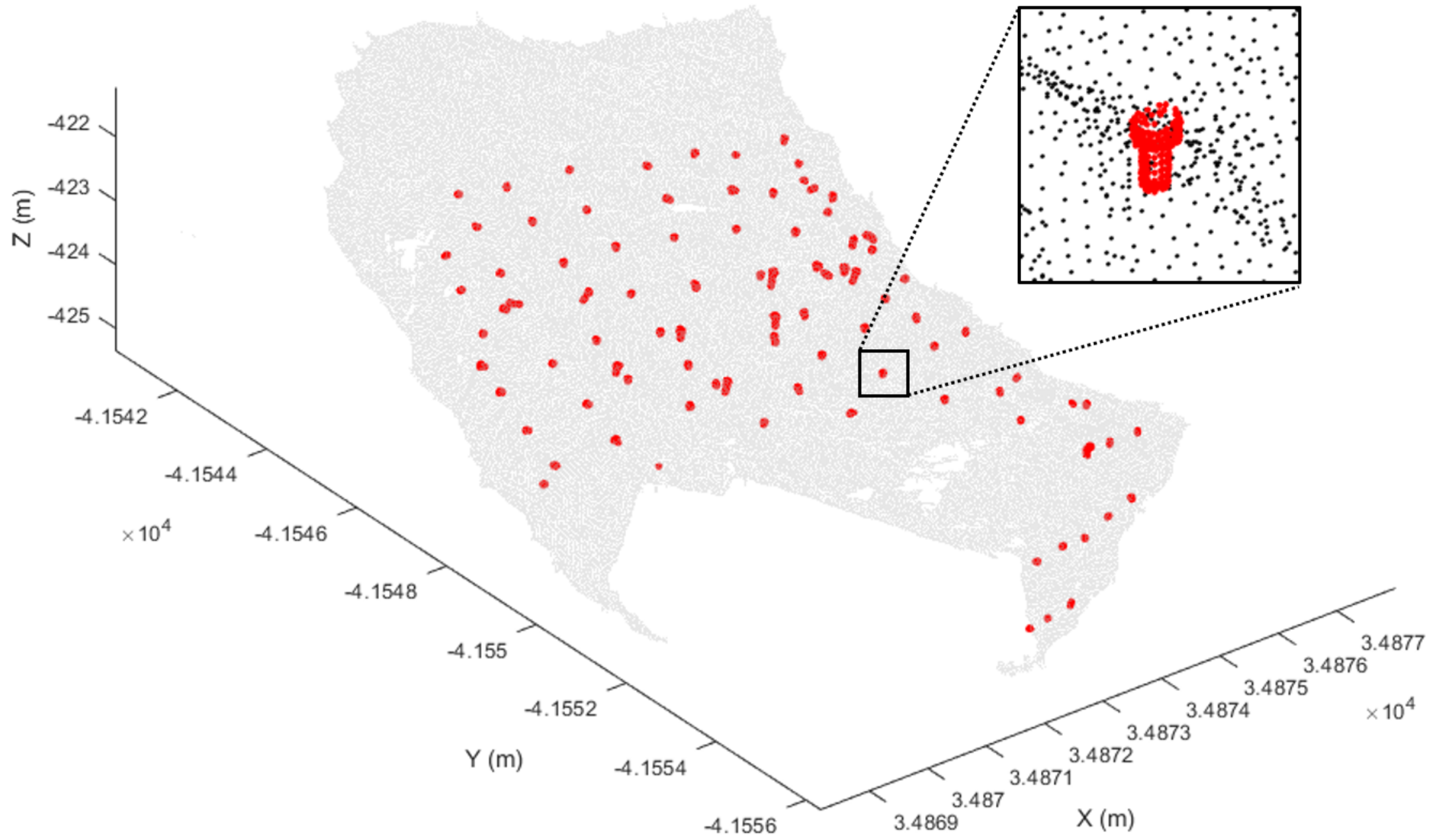

***Figure 6.*** *Rock bolt identification visual results on Scan 1, with a zoomed-in snippet showing an example identified rock bolt.*

The quantitative performance of the rock bolt identification on Scan 1 is summarised in Table 4. At the object level, the detection outcomes are expressed in terms of true positives (TP), false negatives (FN), and false positives (FP). In this study, true positives denote bolts that were present and correctly identified, false negatives denote actual bolts that were not detected by the model, and false positives denote detections incorrectly assigned to bolt protrusions despite not corresponding to real bolts. From a total manually annotated ground truth count of 89 bolts, the approach produced 83 true positives, 4 false positives, and 6 false negatives, indicating that the majority of installed bolts were correctly identified, with only a limited number of missed or spurious detections. The observed false positives are primarily attributed to localised high-curvature or approximately cylindrical non-bolt structures or residual noise retained, which can exhibit geometric characteristics similar to bolt protrusions. Conversely, the false negatives are mainly associated with partial occlusion and reduced point visibility, particularly towards the far end of the scan, where point sparsity and shadowing effects are inherent limitations of the scanning technique.

***Table 4.*** *Rock bolt identification performance validation against ground truth.*

| **Detection Counts** | | **Performance Metrics** | |
|---|---|---|---|
| **Bolt Parameter** | **Value** | **Bolt Parameter** | **Value** |
| Ground truth (GT) | 89 | Classification precision | 93.3% |

| True positives (TP) | 83 | Classification recall | 95.4% |
|---|---|---|---|
| False positives (FP) | 4 | Segmentation IoU | 91.2% |
| False negatives (FN) | 6 | Segmentation precision | 91.9% |

To further evaluate object-level performance, identification precision and identification recall were computed. Identification precision quantifies the proportion of detected bolts that correspond to actual bolt protrusions, thereby reflecting the reliability of positive detections. The obtained precision of 93.3% indicates that the identified bolt instances are highly reliable. Identification recall quantifies the proportion of ground truth bolts successfully recovered by the model, thereby reflecting the completeness of the identification process. The obtained recall of 95.4% indicates that nearly all visible bolts within the scan were detected.

In addition to object-level evaluation, point-level segmentation performance was assessed using segmentation precision and Intersection over Union (IoU). Segmentation precision measures the proportion of points classified as bolt points that truly belong to bolt protrusions, and therefore reflects the degree of contamination from surrounding rock mass or background points within the predicted bolt regions. The obtained segmentation precision of 91.9% indicates that the predicted bolt segments contain minimal inclusion of non-bolt points. IoU measures the overlap between the predicted and reference bolt point sets relative to their union, and provides a stricter assessment of segmentation quality by simultaneously penalising over-segmentation and under-segmentation. The IoU value of 91.2% demonstrates strong correspondence between the predicted bolt regions and the manually annotated reference data.

Taken together, these results demonstrate strong performance at both the object and point levels. The high identification precision and recall indicate that the method is both reliable and nearly exhaustive in detecting visible bolt instances, while the high segmentation precision and IoU confirm that the spatial extent of the bolt protrusions is delineated with high fidelity. These results support the suitability of the proposed approach for robust rock bolt identification in underground mine point clouds and for subsequent integration within the integrated visualisation framework.

### 5.3. Integrated Rock Support Visualisation

The integrated 3D visualisation presented in Figure 7 provides a spatially explicit and integrated representation of the relationship between the characterised discontinuity planes and the installed rock bolt support system. This visualisation is generated by directly combining the discontinuity plane fitting and rock bolt orientation estimation methodologies to form an explicit and integrated representation of the rock support design. Fitted discontinuity planes are colour-coded by set membership and overlaid with rock bolt vectors represented as distinct red cylindrical vectors, enabling geometric assessment of support placement and orientation relative to the structural fabric of the rock mass. At the scale of the full Scan 1 dataset (Figure 7a), the visualisation captures both the distribution of bolts and their orientations across the excavation and the spatial arrangement of the mapped structural planes. This allows mining personnel to identify areas of consistent bolt alignment, variable bolt orientation, and potentially unsupported structural features to be identified within a single 3D framework. The zoomed-in roof-wall section (Figure 7b) from this visualisation provides more detailed local-scale insight, allowing individual bolts to be assessed relative to nearby intersecting discontinuity planes. Several bolts are observed to intersect or orient favourably against major structural planes, suggesting effective reinforcement. However, in the visualisation, it can also be seen that certain structural planes have limited or no direct bolt interaction, which may indicate zones requiring further assessment for potential rock failure. Furthermore, bolts properly aligned and intersecting dominant discontinuity planes are likely to provide greater resistance against potential block displacement, whereas bolts poorly aligned with the relevant structural planes may contribute less effectively to rock mass stabilisation, which can be visually assessed in this representation. Overall, this integrated visualisation provides a practical qualitative tool for mining and geotechnical engineers to assess support effectiveness within the true excavation geometry. By combining structural mapping and bolt orientation information in a single 3D representation, the approach supports more rapid, site-specific interpretation while reducing reliance on manual underground inspection under low-light and potentially hazardous conditions.

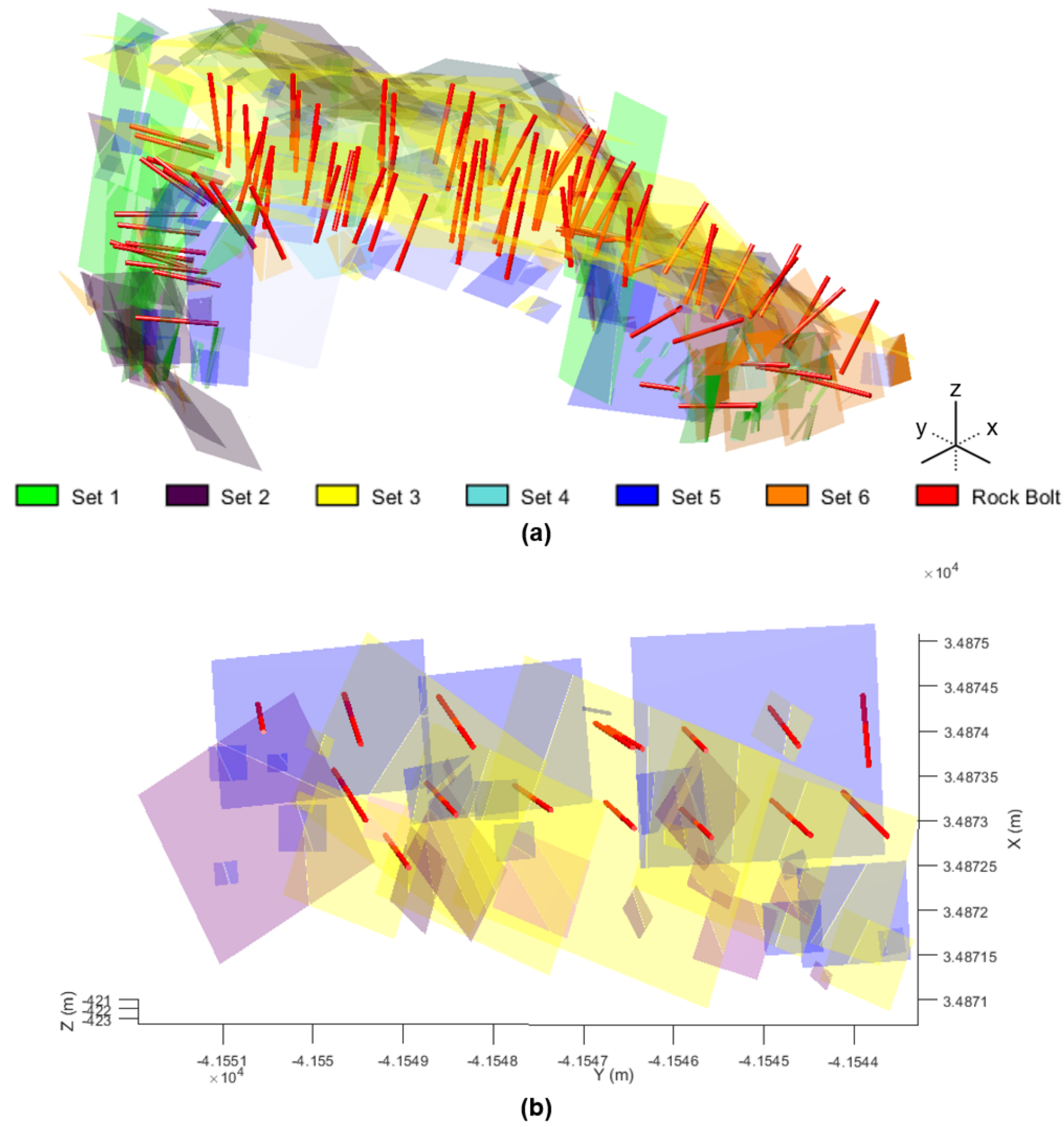

***Figure 7.*** *Integrated 3D visualisation of fitted discontinuity planes and rock bolt vectors for (a) the full Scan 1 point cloud and (b) the selected zoomed-in roof wall section.*

Figure 8 presents a qualitative and quantitative assessment of rock bolt performance using two key quality metrics: exposed protrusion length and angular deviation from the local roof normal. In Figure 8(a), bolt vectors are colour-coded by their visible protrusion length, providing a spatial indication of installation depth and consistency across the excavation. Most bolts fall within a relatively consistent range, with a median protrusion length of ~12 cm and a low standard deviation of ~2 cm. However, several bolts exhibit noticeably greater protrusion lengths, as indicated by the upper-range colours and corresponding outliers in the box-and-whisker plot. Figure 8(b) presents bolt vectors colour-coded by their angular deviation from the local roof normal, providing an indication of bolt alignment quality. The results show a wider spread in deviation values, with a median of ~14° and a standard deviation of ~9°, indicating greater variability in bolt orientation. Regions containing bolts with larger angular deviations may reflect localised installation inconsistencies or deformation effects. These metrics provide indicators of installation quality. In addition to providing quantitative values, the metrics are qualitatively encoded into the point cloud using a colour scale, enabling mine personnel to localise bolts of concern for assessment and planning. Excessive protrusion length may reflect improper embedment, incomplete installation, or alternatively the influence of shear forces within the rock mass pushing the bolt outward. Similarly, large deviations from the local roof normal may indicate suboptimal bolt alignment, reduced support effectiveness, or deformation of bolts due to stress-induced bending. Both metrics are geotechnically significant, as they offer measurable indicators of support performance and potential instability mechanisms. The combined spatial and statistical representation enables engineers to identify potential anomalous regions, assess the consistency of installation practices, and pinpoint areas where support performance may be compromised. This provides a practical and data-driven basis for targeted inspection, maintenance, and optimisation of bolting strategies.

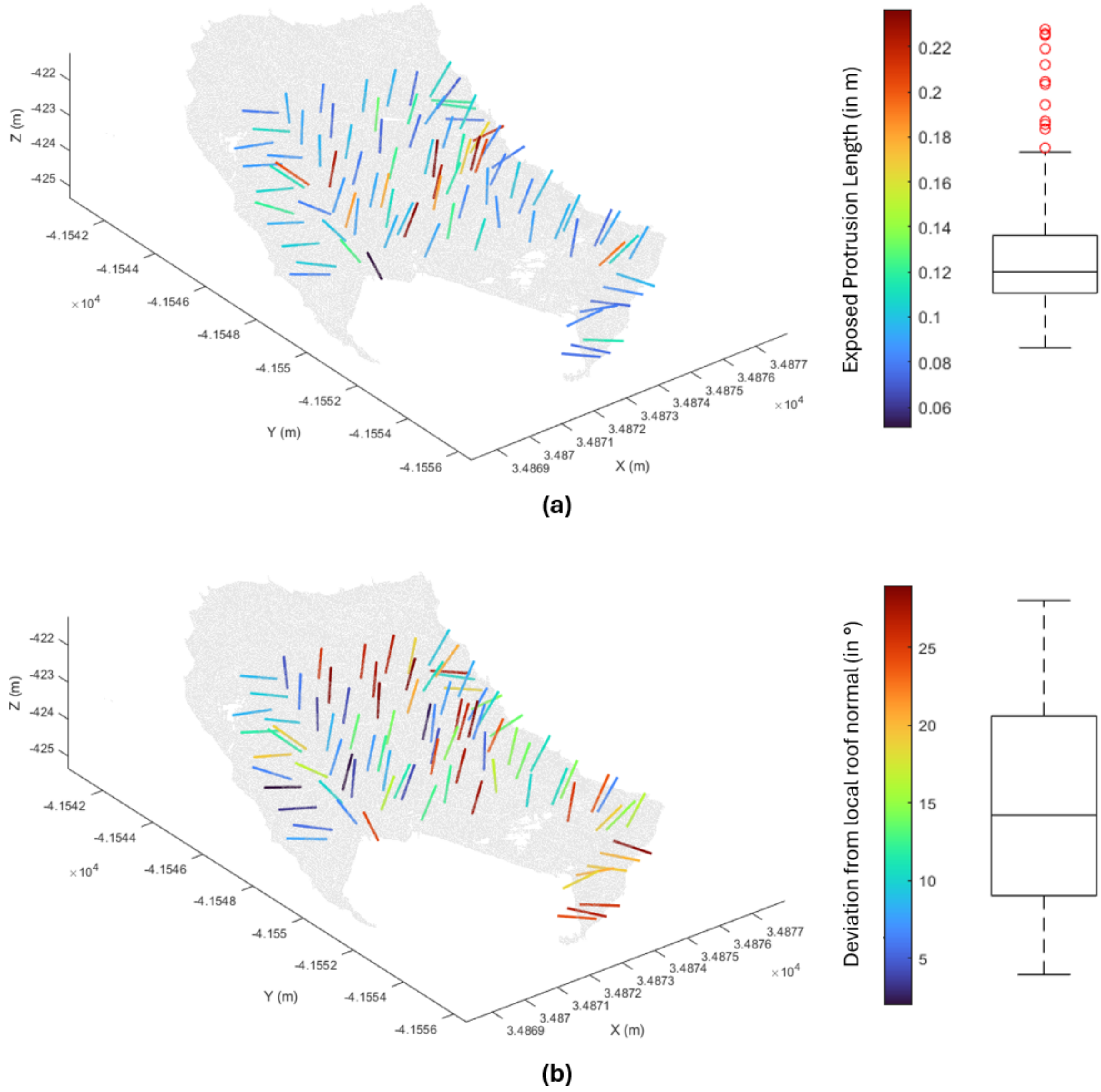

***Figure 8.*** *Rock bolt vectors colour-coded by quality assessment metrics: (a) exposed protrusion length and (b) angular deviation from the local roof normal. Corresponding box-and-whisker plots show the statistical distribution and spread of each metric within the colour spectrum.*

Complementing the spatial 3D visualisation, the stereographic analysis presented in Figure 9 provides a quantitative comparison of structural orientations and rock bolt alignment in orientation space. A total of 196 discontinuity plane poles and 89 rock bolt vectors are projected onto the stereonet, enabling direct evaluation of the relationship between the discontinuity sets and installed support. The discontinuity poles (Figure 9a) form clearly defined clusters, with each set exhibiting tight concentration around its mean orientation and minimal maximum dispersion of less than 15°, indicating robust and consistent discontinuity set identification. To represent the extent of each set, representative cones with a radius of 15° are defined in the stereonet space, capturing the spread of orientations within each discontinuity set. The rock bolt orientations (Figure 9b) show a comparatively broader distribution, reflecting variability in installation alignment and local geometric constraints. When integrated with the discontinuity set envelopes (Figure 9c), the stereonet allows for direct assessment of the geometric effectiveness of rock support design in orientation space. The results indicate that a total of 7 bolts fall outside the defined discontinuity set spreads, suggesting that they are not effectively oriented to support any identified structural set. Furthermore, discontinuity Sets 2 and 5 show no overlap with bolt orientations, indicating a complete lack of direct structural support for these sets. In contrast, Set 3 exhibits the highest degree of overlap with bolt orientations, suggesting it is the most effectively supported set. The remaining sets display partial overlap, indicating some level of support, albeit not consistently across all structural features. This stereographic representation provides a complementary perspective to the 3D visualisation by quantifying the geometric alignment between structural discontinuities and support elements. The ability to assess support coverage in orientation space enables identification of unsupported or under-supported discontinuity sets, offering critical insight for evaluating the adequacy of current bolting strategies and guiding targeted improvements in support design.

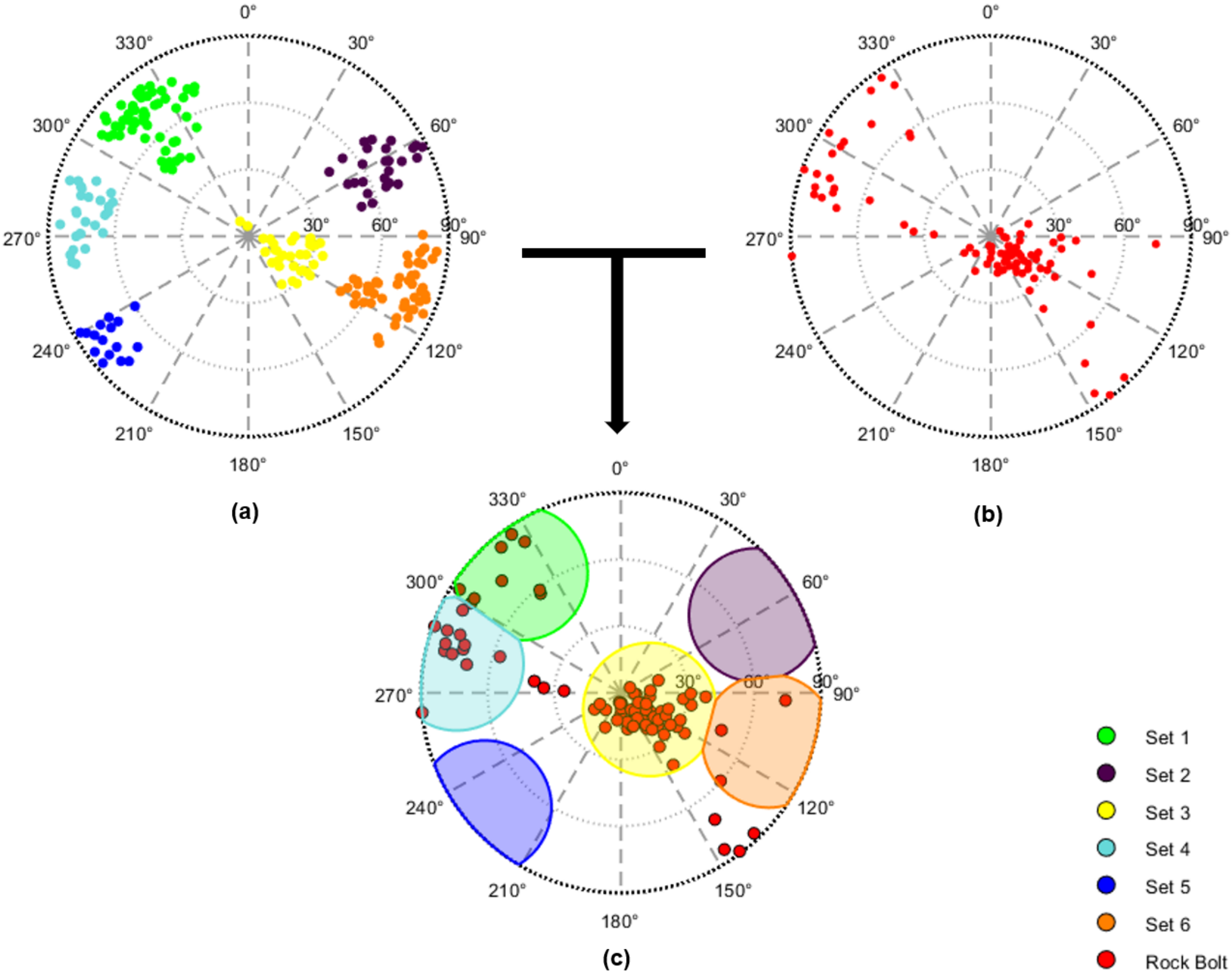

***Figure 9.*** *Stereographic analysis of structural and support orientations. (a) Discontinuity plane poles colour-coded by discontinuity set, (b) rock bolt vector orientations, and (c) integrated stereonet visualisation showing bolt orientations overlaid on discontinuity set orientation envelopes for assessing bolt alignment and support effectiveness.*

### 5.4. Processing Times

The overall processing time for the proposed workflow is summarised in Table 5, which presents a stepwise breakdown of execution time across all major stages. The test dataset (Scan 1) consists of approximately 22 million points, which is reduced to ~700,000 points through voxel-based downsampling. This reduction significantly improves computational efficiency while preserving the essential geometric characteristics required for subsequent analysis. Several design considerations were implemented to optimise performance across the workflow. Even the feature descriptors, particularly eigenvalue-based geometric attributes, are computed once and reused across both structure mapping and rock bolt identification pipelines. This avoids redundant computations and enables efficient reuse and repurposing of intermediate results. In addition, geometric filtering steps applied in both workflows eliminate insignificant points, improving both computational efficiency and analytical accuracy. The impact of filtering is particularly pronounced in the rock bolt identification stage, where more than 90% of background points are removed prior to semantic segmentation. This targeted reduction ensures that only regions of interest are processed by the deep learning model, resulting in substantially faster inference times without compromising detection performance. The semantic segmentation model was trained using the Adam optimiser with an initial learning rate of 0.001, momentum of 0.1, and batch size of 16. A momentum decay of 0.5 was applied every 16 steps, with convergence achieved after 32 epochs (~8 hours of training time). All computations were performed on a 64-bit Windows system equipped with an Intel W-2245 CPU, 128 GB RAM, and an NVIDIA T1000 GPU (8 GB VRAM, 896 CUDA cores). The total execution time for Scan 1 is approximately 348 seconds (~5.8 minutes), demonstrating that the proposed framework is computationally efficient for medium-scale datasets. This level of performance is achieved while maintaining high accuracy, as demonstrated in the preceding validation results, making the method suitable for practical deployment in mining environments for planning, design, and quality assessment. The execution time analysis further indicates that eigenvalue and normal computation constitute the primary computational bottleneck, accounting for the majority of total processing time. This is expected due to the per-point neighbourhood-based PCA operations required for geometric characterisation. All subsequent stages, including clustering, deep learning inference, and visualisation, incur comparatively lower computational overhead. This highlights that once local geometric descriptors are established, the overall framework operates efficiently and scales well for integrated structural and support analysis.

***Table 5.*** *Overall execution time for integrated rock support visualisation on Scan 1.*

| Type | Process | Time Taken (s) |
|---|---|---|
| Pre-processing | Statistical outlier removal filter | 5.36 |
| | Voxel downsampling | <1.00 |
| | Cloth simulation filter | <1.00 |
| | Eigenvalue and point normals | 211.87 |
| Structure mapping | Geometric filtering | <1.00 |
| | Discontinuity set clustering | 65.49 |
| Rock bolt identification | Geometric Filtering | 9.20 |
| | Semantic Segmentation | 47.51 |
| Integrated rock support visualisation | Discontinuity plane fitting | 2.89 |
| | Rock bolt orientation estimation | 3.15 |
| | Integrated 3D visualisation | <1.00 |
| | Stereographic Analysis | <1.00 |
| | **Total Time** | **≈ 348 s (5.8 minutes)** |

## 6. Limitations and Future Research

This study presents a proof of concept for integrated rock support visualisation, demonstrating its potential as an automated and efficient framework for routine geotechnical monitoring, geometric support quality assessment, and mine planning. However, despite the effectiveness of the proposed framework for geometric assessment and visualisation of rock support interaction with structural features, the approach remains inherently limited in evaluating the full mechanical effectiveness of installed support systems. Since the framework is based entirely on remote sensing and surface exposed 3D point cloud information, critical in-situ parameters governing support behaviour, such as mechanical anchorage condition, pretension, load transfer, corrosion state, and internal rock mass interaction, cannot be directly assessed. In addition, the estimated bolt orientations are derived only from the visible protruding sections of the support elements captured within the point cloud. As the embedded portion of the bolt within the rock mass is not observable using laser scanning systems, the estimated orientation represents an approximation of the exposed bolt axis rather than the complete in-situ bolt trajectory. Similarly, the discontinuity characterisation process is constrained to exposed excavation surfaces and does not directly capture true persistence, concealed geologically significant discontinuities, or structural continuity beyond the scanned rock face. Consequently, the framework is best interpreted as a geometric and spatial assessment tool for support quality assurance and quality control analysis rather than a direct measure of mechanical rock support capacity or long-term mechanical support performance. Nevertheless, results establish a robust foundation upon which several promising research directions can be pursued to further advance the capability, applicability, and practical deployment of the proposed framework in operational mining environments.

A natural and impactful extension of the current work is the implementation of the integrated visualisation framework within immersive extended reality environments, including holographic display platforms and virtual reality systems. Current 2D and 3D screen-based visualisations, while informative, do not fully exploit the spatial and structural richness of the point cloud data. Deploying the framework within interactive platforms such as holographic tables, head-mounted virtual reality devices, or virtual reality projectors would enable mine engineers and geotechnical practitioners to navigate, interrogate, and interact with the full 3D structural model from off-site control rooms during planning and assessment activities. Development of interactive modules, for instance, using game engine frameworks such as Unity, would allow users to manipulate bolt vector representations, reposition structural elements, design and evaluate support layouts in digital space, and simulate alternative bolting configurations prior to physical installation or planning post it. Such a capability would transform the current visualisation output into a fully functional digital twin-based mine planning tool, enabling immersive, collaborative, and data-driven support design from remote locations and representing a significant step towards next-generation underground mine planning infrastructure.

The current framework operates on single-epoch point cloud scans, providing a static snapshot of rock mass structure and installed support. A critical extension would be the incorporation of multi-temporal point cloud data to enable change detection and time-series monitoring of both structural features and rock bolt conditions. By comparing successive scans acquired at regular intervals, ideally through automated drone-based scanning systems capable of routine data collection without human intervention, the framework could detect progressive changes in discontinuity orientations, the emergence of new structural features, and deviations in bolt orientations or protrusion lengths indicative of structural movement or bolt degradation. Integrating predictive analytics with this temporal data stream would enable the development of automated predictive maintenance systems capable of identifying deteriorating trends and forecasting potential failure mechanisms before critical thresholds are reached. Such a system would provide a continuous, data-driven early warning capability for ground control, substantially improving safety outcomes in dynamic underground environments.

The current framework assesses rock support exclusively from geometric point cloud data, providing a surface-level spatial understanding of installed bolt orientations and positions. While sufficient for orientation-based quality assessment, this does not capture the mechanical state of the support system, specifically the forces, deformations, and load distributions acting on individual bolts in response to rock mass movement and stress redistribution. A natural extension would be the integration of smart bolt technology, whereby bolts are instrumented with embedded sensors, including strain gauges, load cells, and accelerometers, to measure axial load, shear force, and deformation in real time. Coupling these measurements with the geometric bolt representations produced by the proposed framework would enable each bolt in the 3D visualisation and stereonet to be attributed with live mechanical performance data alongside its geometric descriptors. Further integration of micro-seismic monitoring data would provide additional context on dynamic loading conditions and stress redistribution in the surrounding rock mass. Together, this multi-modal fusion would transition the framework from geometric support assessment to a comprehensive mechanical performance monitoring system, providing a substantially richer basis for predictive maintenance and ground control decision-making in operational underground mining environments.

Although the current study shares several common computational modules across the structure mapping and rock bolt identification pipelines for efficiency, each workflow retains dedicated processing stages tailored to the specific geometric properties of its target features. A significant long-term research direction is the development of a unified deep learning architecture capable of simultaneously identifying and classifying all geotechnically relevant structural and operational features present in underground mine point clouds in one go using a single end-to-end framework. Realising this vision would require the assembly of a large-scale, diverse annotated dataset spanning multiple mine types, geological settings, scanner configurations, and excavation geometries. Establishing such a repository as a community benchmark would address the critical data scarcity that currently limits generalisation in data-driven approaches for underground mine point cloud understanding and would provide the training foundation necessary to develop robust, universally deployable models for automated geotechnical characterisation at scale.

## 7. Conclusions

This study presented an automated framework for integrated rock support visualisation using 3D point cloud data from underground metal mining environments. The framework integrated structure mapping and rock bolt identification into a unified workflow, enabling direct correlation between discontinuity planes and installed rock bolt orientations. By incorporating discontinuity plane fitting and rock bolt orientation estimation, the method generated explicit geometric representations that formed the basis of both integrated 3D visualisation and stereographic analysis. The results demonstrated that the proposed approach is capable of accurately characterising discontinuity sets and reliably identifying rock bolts in real-world underground mine datasets. The structure mapping method achieved orientation errors within the range of manual interpretation variability, while the rock bolt identification pipeline demonstrated high precision and recall at both object and point levels. The integrated visualisation framework enabled clear spatial and orientation-based assessment of the geometric effectiveness of the installed rock support, allowing identification of well-supported structural features, misaligned bolts, and unsupported discontinuity sets. In addition to accuracy, the framework was computationally efficient. Through the reuse of shared preprocessing steps and feature descriptors, redundant computations were minimised, allowing the complete workflow to process medium-scale point clouds efficiently. Overall, the proposed framework provided a practical, automated, and surface-faithful solution for assessing rock support effectiveness directly from point cloud data, representing a significant step towards integrated rock support visualisation.

**Acknowledgements**

Dibyayan Patra acknowledges the financial support provided by the University International Postgraduate Award (UIPA) from the University of New South Wales for this research. The authors also acknowledge the facilities and resources provided by the Laboratory for Imaging of the Mine Environment at the University of New South Wales, Sydney, for this research.

**Author contributions**

CRediT: **Dibyayan Patra:** Writing – original draft, Data curation, Investigation, Conceptualisation, Methodology, Software, Formal analysis, Visualisation, Validation. **Simit Raval:** Writing – review and editing, Project administration, Data curation, Resources, Supervision. **Pasindu Ranasinghe:** Writing – review and editing, Formal Analysis, Validation. **Bikram Banerjee:** Writing – review and editing, Validation, Supervision. **Ismet Canbulat:** Writing – review and editing, Supervision.

**Declaration of generative AI and AI-assisted technologies in the writing process**

During the preparation of this work, the authors used OpenAI's ChatGPT (version GPT-5.5) to improve the language and readability of the manuscript. After using this tool, the authors reviewed and edited the content as needed and take full responsibility for the content of the published article.

**Disclosure statement**

The authors declare that they have no known competing financial interests or personal relationships that could have appeared to influence the work reported in this paper.

**Data availability statement**

Raw point cloud data cannot be made available due to non-disclosure agreements with mine site. Other research data will be made available upon reasonable request.

**Funding**

No funding was received.

## REFERENCES

[1] C. C. Li, "Principles of rockbolting design," *Journal of Rock Mechanics and Geotechnical Engineering,* vol. 9, no. 3, pp. 396-414, 2017, doi: 10.1016/j.jrmge.2017.04.002.

[2] B. H. G. Brady and E. T. Brown, "Rock support and reinforcement," in *Rock Mechanics for underground mining: Third edition*, B. H. G. Brady and E. T. Brown Eds. Dordrecht: Springer Netherlands, 2006, pp. 312-346.

[3] Y. Cai, T. Esaki, and Y. Jiang, "A rock bolt and rock mass interaction model," *International Journal of Rock Mechanics and Mining Sciences,* vol. 41, no. 7, pp. 1055-1067, 2004, doi: 10.1016/j.ijrmms.2004.04.005.

[4] M. Ghorbani, K. Shahriar, M. Sharifzadeh, and R. Masoudi, "A critical review on the developments of rock support systems in high stress ground conditions," *International Journal*

*of Mining Science and Technology,* vol. 30, no. 5, pp. 555-572, 2020, doi: 10.1016/j.ijmst.2020.06.002.
[5] K. Peter *et al.*, "An Overview of the Use of Rockbolts as Support Tools in Mining Operations," *Geotechnical and Geological Engineering,* vol. 40, no. 4, pp. 1637-1661, 2021, doi: 10.1007/s10706-021-02005-5.
[6] P. Singh, H. Jang, and A. J. S. S. Spearing, "Improving the Numerical Modelling of In-Situ Rock Bolts Using Axial and Bending Strain Data from Instrumented Bolts," *Geotechnical and Geological Engineering,* vol. 40, no. 5, pp. 2631-2655, 2022, doi: 10.1007/s10706-022-02051-7.
[7] Y. Xing, P. H. S. W. Kulatilake, and L. A. Sandbak, "Effect of rock mass and discontinuity mechanical properties and delayed rock supporting on tunnel stability in an underground mine," *Eng Geol,* vol. 238, pp. 62-75, 2018, doi: 10.1016/j.enggeo.2018.03.010.
[8] C. Pu, J. Zhan, W. Zhang, and J. Peng, "Characterization and clustering of rock discontinuity sets: A review," *Journal of Rock Mechanics and Geotechnical Engineering,* vol. 17, no. 2, pp. 1240-1262, 2025, doi: 10.1016/j.jrmge.2024.03.041.
[9] S. D. Priest, "Discontinuities and rock strength," in *Discontinuity Analysis for Rock Engineering*, 1993, ch. Chapter 9, pp. 259-299.
[10] R. Battulwar, M. Zare-Naghadehi, E. Emami, and J. Sattarvand, "A state-of-the-art review of automated extraction of rock mass discontinuity characteristics using three-dimensional surface models," *Journal of Rock Mechanics and Geotechnical Engineering,* vol. 13, no. 4, pp. 920-936, 2021, doi: 10.1016/j.jrmge.2021.01.008.
[11] H. Daghigh, D. D. Tannant, V. Daghigh, D. D. Lichti, and R. Lindenbergh, "A critical review of discontinuity plane extraction from 3D point cloud data of rock mass surfaces," *Computers & Geosciences,* vol. 169, 2022, doi: 10.1016/j.cageo.2022.105241.
[12] S. K. Singh, B. P. Banerjee, and S. Raval, "A review of laser scanning for geological and geotechnical applications in underground mining," *International Journal of Mining Science and Technology,* vol. 33, no. 2, pp. 133-154, 2023/02/01/ 2023, doi: 10.1016/j.ijmst.2022.09.022.
[13] D. Zhao *et al.*, "Data-driven intelligent prediction of TBM surrounding rock and personalized evaluation of disaster-inducing factors," *Tunn Undergr Sp Tech,* vol. 148, 2024, doi: 10.1016/j.tust.2024.105768.
[14] S. Adhikari, K. K. Panthi, and C. B. Basnet, "Subjectivity associated to the use of rock mass classification in stability analysis of caverns," *Sci Rep,* vol. 15, no. 1, p. 26256, Jul 19 2025, doi: 10.1038/s41598-025-05055-4.
[15] J. Chen, Q. Fang, D. Zhang, and H. Huang, "A critical review of automated extraction of rock mass parameters using 3D point cloud data," *Intelligent Transportation Infrastructure,* vol. 2, 2023, doi: 10.1093/iti/liad005.
[16] S. Fekete, M. Diederichs, and M. Lato, "Geotechnical and operational applications for 3-dimensional laser scanning in drill and blast tunnels," *Tunn Undergr Sp Tech,* vol. 25, no. 5, pp. 614-628, 2010, doi: 10.1016/j.tust.2010.04.008.
[17] A. Riquelme, M. Cano, R. Tomás, and A. Abellán, "Identification of Rock Slope Discontinuity Sets from Laser Scanner and Photogrammetric Point Clouds: A Comparative Analysis," *Procedia Engineering,* vol. 191, pp. 838-845, 2017, doi: 10.1016/j.proeng.2017.05.251.
[18] Y. Li, Q. Wang, J. Chen, L. Xu, and S. Song, "K-means Algorithm Based on Particle Swarm Optimization for the Identification of Rock Discontinuity Sets," *Rock Mechanics and Rock Engineering,* vol. 48, no. 1, pp. 375-385, 2014, doi: 10.1007/s00603-014-0569-x.
[19] J. Guo, S. Liu, P. Zhang, L. Wu, W. Zhou, and Y. Yu, "Towards semi-automatic rock mass discontinuity orientation and set analysis from 3D point clouds," *Computers & Geosciences,* vol. 103, pp. 164-172, 2017, doi: 10.1016/j.cageo.2017.03.017.
[20] J.-w. Zhou, J.-l. Chen, and H.-b. Li, "An optimized fuzzy K-means clustering method for automated rock discontinuities extraction from point clouds," *International Journal of Rock Mechanics and Mining Sciences,* vol. 173, 2024, doi: 10.1016/j.ijrmms.2023.105627.
[21] E. Mammoliti, F. Di Stefano, D. Fronzi, A. Mancini, E. S. Malinverni, and A. Tazioli, "A Machine Learning Approach to Extract Rock Mass Discontinuity Orientation and Spacing, from Laser Scanner Point Clouds," *Remote Sensing,* vol. 14, no. 10, 2022, doi: 10.3390/rs14102365.
[22] J. Chen, H. Huang, M. Zhou, and K. Chaiyasarn, "Towards semi-automatic discontinuity characterization in rock tunnel faces using 3D point clouds," *Eng Geol,* vol. 291, 2021, doi: 10.1016/j.enggeo.2021.106232.
[23] C.-C. Chiu and C.-Y. Liu, "Development of a computer program from photogrammetry for assisting Q-system rating," *International Journal of Rock Mechanics and Mining Sciences,* vol. 170, 2023, doi: 10.1016/j.ijrmms.2023.105499.

[24] A. M. Ikotun, F. Habyarimana, and A. E. Ezugwu, "Cluster validity indices for automatic clustering: A comprehensive review," *Heliyon,* vol. 11, no. 2, p. e41953, Jan 30 2025, doi: 10.1016/j.heliyon.2025.e41953.

[25] A. J. Riquelme, A. Abellán, R. Tomás, and M. Jaboyedoff, "A new approach for semi-automatic rock mass joints recognition from 3D point clouds," *Computers & Geosciences,* vol. 68, pp. 38-52, 2014, doi: 10.1016/j.cageo.2014.03.014.

[26] A. J. Riquelme, A. Abellán, and R. Tomás, "Discontinuity spacing analysis in rock masses using 3D point clouds," *Eng Geol,* vol. 195, pp. 185-195, 2015, doi: 10.1016/j.enggeo.2015.06.009.

[27] A. Riquelme, R. Tomás, M. Cano, J. L. Pastor, and A. Abellán, "Automatic Mapping of Discontinuity Persistence on Rock Masses Using 3D Point Clouds," *Rock Mechanics and Rock Engineering,* vol. 51, no. 10, pp. 3005-3028, 2018, doi: 10.1007/s00603-018-1519-9.

[28] D. Patra, C. Baylis, P. Ranasinghe, B. Banerjee, and S. Raval, "A UAV Laser Scanning Technique for Automated Mapping of In-Stope Structural Discontinuity Sets in Underground Mines," presented at the IGARSS 2025 - 2025 IEEE International Geoscience and Remote Sensing Symposium, 2025.

[29] S. K. Singh, S. Raval, and B. P. Banerjee, "Automated structural discontinuity mapping in a rock face occluded by vegetation using mobile laser scanning," *Eng Geol,* vol. 285, 2021, doi: 10.1016/j.enggeo.2021.106040.

[30] S. K. Singh, B. P. Banerjee, M. J. Lato, C. Sammut, and S. Raval, "Automated rock mass discontinuity set characterisation using amplitude and phase decomposition of point cloud data," *International Journal of Rock Mechanics and Mining Sciences,* vol. 152, p. 105072, 2022/04/01/ 2022, doi: 10.1016/j.ijrmms.2022.105072.

[31] X. Wang, L. Zou, X. Shen, Y. Ren, and Y. Qin, "A region-growing approach for automatic outcrop fracture extraction from a three-dimensional point cloud," *Computers & Geosciences,* vol. 99, pp. 100-106, 2017, doi: 10.1016/j.cageo.2016.11.002.

[32] Y. Ge *et al.*, "Automated measurements of discontinuity geometric properties from a 3D-point cloud based on a modified region growing algorithm," *Eng Geol,* vol. 242, pp. 44-54, 2018, doi: 10.1016/j.enggeo.2018.05.007.

[33] N. Chen, X. Wu, H. Xiao, C. Yao, and Y. Cheng, "Semi-automatic recognition of rock mass discontinuity based on 3D point clouds," *Discover Applied Sciences,* vol. 6, no. 5, 2024, doi: 10.1007/s42452-024-05876-4.

[34] V. A. Puligandla and S. Loncaric, "A Supervoxel Segmentation Method With Adaptive Centroid Initialization for Point Clouds," *IEEE Access,* vol. 10, pp. 98525-98534, 2022, doi: 10.1109/access.2022.3206387.

[35] W. Sun, J. Wang, Y. Yang, and F. Jin, "Rock Mass Discontinuity Extraction Method Based on Multiresolution Supervoxel Segmentation of Point Cloud," *IEEE Journal of Selected Topics in Applied Earth Observations and Remote Sensing,* vol. 14, pp. 8436-8446, 2021, doi: 10.1109/jstars.2021.3104845.

[36] T. J. B. Dewez, D. Girardeau-Montaut, C. Allanic, and J. Rohmer, "Facets : A Cloudcompare Plugin to Extract Geological Planes from Unstructured 3d Point Clouds," *The International Archives of the Photogrammetry, Remote Sensing and Spatial Information Sciences,* vol. XLI-B5, pp. 799-804, 2016, doi: 10.5194/isprs-archives-XLI-B5-799-2016.

[37] R. Battulwar *et al.*, "Utilizing Deep Learning for the Automated Extraction of Rock Mass Features from Point Clouds," *Geotechnical and Geological Engineering,* vol. 42, no. 7, pp. 6179-6194, 2024, doi: 10.1007/s10706-024-02886-2.

[38] Q. Chen, Y. Ge, and H. Tang, "Rock discontinuities characterization from large-scale point clouds using a point-based deep learning method," *Eng Geol,* vol. 337, 2024, doi: 10.1016/j.enggeo.2024.107585.

[39] J. Sun *et al.*, "A robust deep learning approach for rock discontinuity identification from large scale 3D point clouds," *Sci Rep,* vol. 16, no. 1, p. 1654, Dec 16 2025, doi: 10.1038/s41598-025-31137-4.

[40] Y. Ge, H. Wang, G. Liu, Q. Chen, and H. Tang, "Automated Identification of Rock Discontinuities from 3D Point Clouds Using a Convolutional Neural Network," *Rock Mechanics and Rock Engineering,* vol. 58, no. 3, pp. 3683-3700, 2025, doi: 10.1007/s00603-024-04351-1.

[41] G. Lu *et al.*, "Identification of rock mass discontinuity from 3D point clouds using improved fuzzy C-means and convolutional neural network," *Bulletin of Engineering Geology and the Environment,* vol. 83, no. 5, 2024, doi: 10.1007/s10064-024-03658-1.

[42] Y. Ge, B. Cao, and H. Tang, "Rock Discontinuities Identification from 3D Point Clouds Using Artificial Neural Network," *Rock Mechanics and Rock Engineering,* vol. 55, no. 3, pp. 1705-1720, 2022, doi: 10.1007/s00603-021-02748-w.

[43] J. Wang *et al.*, "An interactive framework integrating segment anything model and structure-from-motion for three-dimensional discontinuity identification in rock masses," *International Journal of Mining Science and Technology,* vol. 35, no. 10, pp. 1695-1711, 2025, doi: 10.1016/j.ijmst.2025.09.005.

[44] S. K. Singh, S. Raval, and B. Banerjee, "Roof bolt identification in underground coal mines from 3D point cloud data using local point descriptors and artificial neural network," *International Journal of Remote Sensing,* vol. 42, no. 1, pp. 367-377, 2021/01/02 2020, doi: 10.1080/2150704x.2020.1809734.

[45] S. K. Singh, S. Raval, and B. Banerjee, "A robust approach to identify roof bolts in 3D point cloud data captured from a mobile laser scanner," *International Journal of Mining Science and Technology,* vol. 31, no. 2, pp. 303-312, 2021/03/01/ 2021, doi: 10.1016/j.ijmst.2021.01.001.

[46] Z. Ren, H. Zhu, L. Zhao, and R. Yuan, "MLS-based recognition and parameter extraction of roadway roof bolts/cables from 3D point clouds," *Sci Rep,* vol. 16, no. 1, p. 6538, Jan 28 2026, doi: 10.1038/s41598-026-37689-3.

[47] S. Saydam, B. Liu, B. Li, W. Zhang, S. K. Singh, and S. Raval, "A Coarse-to-Fine Approach for Rock Bolt Detection From 3D Point Clouds," *IEEE Access,* vol. 9, pp. 148873-148883, 2021, doi: 10.1109/access.2021.3120207.

[48] D. Patra, P. Ranasinghe, B. Banerjee, and S. Raval, "A Deep Learning Approach to Identify Rock Bolts in Complex 3D Point Clouds of Underground Mines Captured Using Mobile Laser Scanners," *Remote Sensing,* vol. 17, no. 15, 2025, doi: 10.3390/rs17152701.

[49] X.-F. Han, Z.-A. Feng, S.-J. Sun, and G.-Q. Xiao, "3D point cloud descriptors: state-of-the-art," *Artificial Intelligence Review,* vol. 56, no. 10, pp. 12033-12083, 2023, doi: 10.1007/s10462-023-10486-4.

[50] I. K. Kazmi, L. You, and J. J. Zhang, "A Survey of 2D and 3D Shape Descriptors," presented at the 2013 10th International Conference Computer Graphics, Imaging and Visualization, 2013.

[51] D. Patra, P. Ranasinghe, B. Banerjee, and S. Raval, "An Investigation on Automated Structure Mapping of Subsurface Cavities Using UAV LiDAR Technology," presented at the 2025 IEEE International Conference on Imaging Systems and Techniques (IST), 2025.

[52] L. McInnes, J. Healy, and S. Astels, "hdbscan: Hierarchical density based clustering," *The Journal of Open Source Software,* vol. 2, no. 11, 2017, doi: 10.21105/joss.00205.